\documentclass{article} 
\usepackage{iclr2027_conference,times}

\usepackage{amsmath,amsfonts,bm}

\def\eqref#1{equation~\ref{#1}}

\def\1{\bm{1}}

\DeclareMathAlphabet{\mathsfit}{\encodingdefault}{\sfdefault}{m}{sl}
\SetMathAlphabet{\mathsfit}{bold}{\encodingdefault}{\sfdefault}{bx}{n}

\usepackage{hyperref}
\usepackage{url}

\usepackage{amsmath}
\usepackage{tabularx}
\usepackage{multirow}
\usepackage{amssymb}
\usepackage{booktabs}
\usepackage{amsfonts}
\usepackage{nicefrac}
\usepackage{microtype}
\usepackage{xcolor}
\usepackage{graphicx}

\title{ReGraph: A Computational Account of\\ Emergent Generalization in the\\ “what” and “where” Dual Visual Streams}

\newcommand{\affilline}[2]{\par{\hangindent=0.6em\hangafter=1 $^{#1}$#2\par}}

\author{%
\parbox[t]{\dimexpr\textwidth-2\tabcolsep\relax}{\raggedright
{\bfseries Hyewon Kang$^{1,3}$,\enspace Jungmin Lee$^{2,3}$,\enspace Ilgyu Lee$^{2,3}$,\enspace Seok-Jun Hong$^{1,2,3,}$\thanks{Corresponding author.}}\vspace{3pt}
\normalfont
\affilline{1}{Department of Intelligent Precision Healthcare Convergence, Sungkyunkwan University, Suwon, Republic~of~Korea}
\affilline{2}{Department of Biomedical Engineering, Sungkyunkwan University, Suwon, Republic~of~Korea}
\affilline{3}{Center for Neuroscience Imaging Research, Institute for Basic Science (IBS), Suwon, Republic~of~Korea}
\vspace{3pt}
\texttt{\{hwkang91,jungmin.lee,ilgyu26\}@g.skku.edu},\enspace \texttt{hongseokjun@skku.edu}}
}

\iclrfinalcopy 

\begin{document}

\maketitle
\lhead{Preprint. Under review.}

\begin{abstract}
Where generalization capacity—the ability to extract context-invariant relational structures—first emerges remains a central question in AI and neuroscience. In the brain, the foundation for this capacity lies upstream of the hippocampus, within the entorhinal cortex, where parallel pathways dissociate relational structure in the medial entorhinal cortex (MEC) from sensory content in the lateral entorhinal cortex (LEC). However, as Eichenbaum argued, this structure–content factorization likely originates earlier, driven by the segregation of the dorsal (‘where’) and ventral (‘what’) visual streams. Supporting this, grid-like firing patterns—a defining cellular signature of MEC context-invariant representations—also appear in preceding neocortical regions (e.g., retrosplenial cortex) along the dorsal pathway. While these findings suggest that relational structures do not arise \emph{de novo} in the hippocampal formation, how such representations are computationally formed along upstream pathways remains unknown. To investigate this \emph{in silico}, we developed \emph{ReGraph}, a dual visual-stream model implemented on a recurrent graph architecture with several biological inductive biases, including retina-driven stream-specialized encoding, dorsal-to-ventral modulation, and dynamic lateral connectivity. Trained on the action benchmark Something-Something V2, ReGraph revealed a pathway-specific emergence of relational mapping: context-invariant codes and grid-like spatial bases uniquely co-emerged along the extended dorsal stream. In contrast, the absence of these representations in single-stream models, unmodulated dual-stream variants, and standard action-recognition baselines (e.g., VideoMAE, SlowFast) implies that our selected inductive biases are indeed critical prerequisites for the emergence of relational structures. Crucially, the grid-like bases turned out not to be mere byproducts of architectural design, as our post-hoc analyses demonstrated that these bases could serve as reusable routing templates for information processing via lateral connectivity. Together, our findings provide a computational account suggesting that generalization may not be a faculty that emerges abruptly within a dedicated region, but a property that already takes shape as sensory information is parsed into factorized streams of hierarchical visual processing.

\end{abstract}

\section{Introduction}
\label{sec:introduction}
A hallmark of human intelligence is the ability to generalize by extracting and reusing abstract task-relational structures transferable across new environments~\citep{tenenbaum2011grow, behrens2018cognitive}. In neuroscience, this ability is often linked to the hippocampal formation, particularly the entorhinal cortex, where a profound functional dissociation emerges:  the lateral entorhinal cortex (LEC) encodes sensory-specific content~\citep{knierim2014functional}, while the medial entorhinal cortex (MEC) extracts generalizable relational structures, represented by grid cells with space-tessellating hexagonal firing patterns~\citep{moser2008place, behrens2018cognitive}.

\begin{figure}[tbp] 
    \centering
    \includegraphics[width=\textwidth, trim=0 0 0 0, clip]{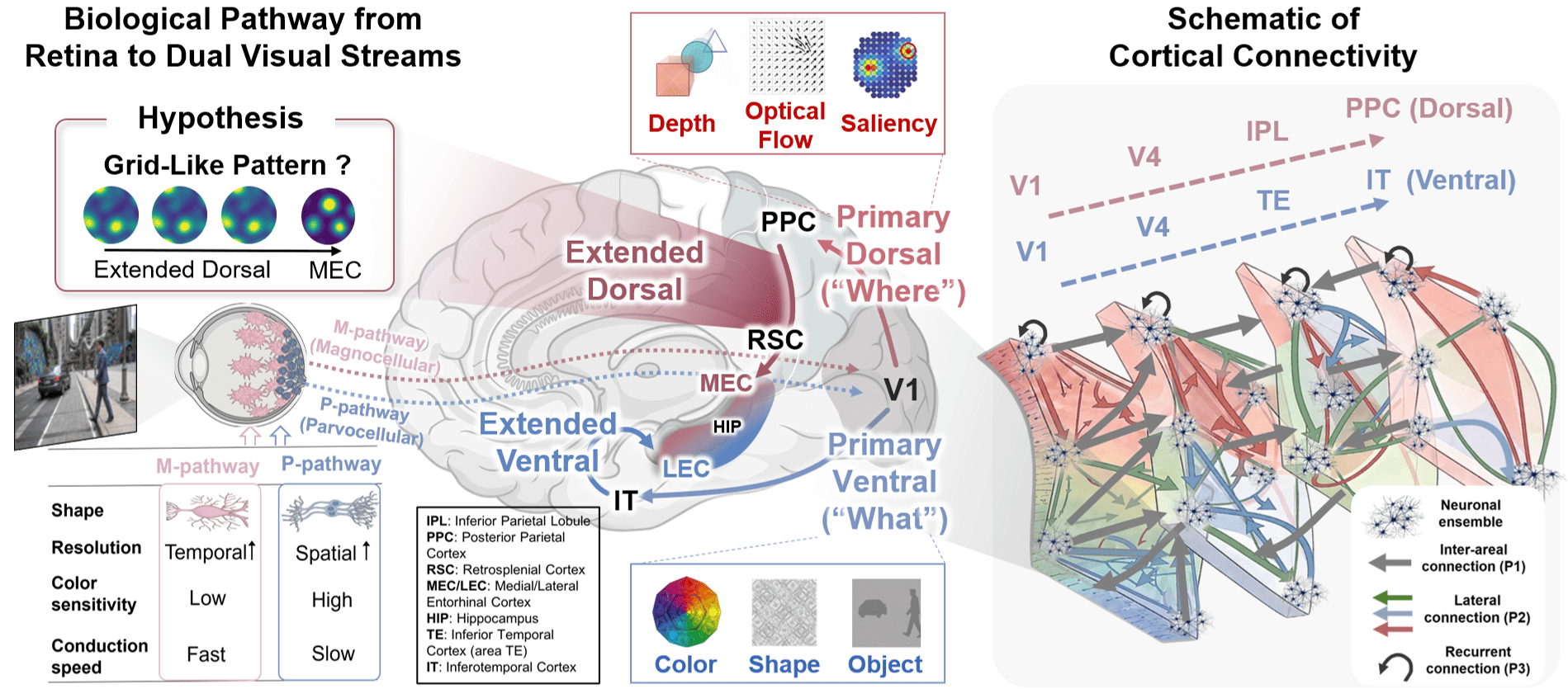}
    \caption{\textbf{From retina to dual visual streams.}
    \textbf{Left}: The retinal M/P dichotomy drives the dorsal and ventral pathways, each comprising primary and extended stages. Grid-like patterns in the extended dorsal pathway motivate a hypothesis that the precursor of relational structures may emerge along this pathway. The primary dorsal/ventral pathways preferentially encode spatiotemporal dynamics and object details, respectively. \textbf{Right:} Schematics of three connectivity principles defined in \S\ref{sec:extended}.}
    \label{fig:overview}
\end{figure}

This MEC/LEC factorization, however, does not originate within the entorhinal cortex. As Eichenbaum and colleagues have long argued, it inherits a deeper cortical segregation: two long-range streams—the `\textit{where}' (dorsal) and `\textit{what}' (ventral) pathways—that provide anatomically segregated inputs to MEC and LEC, respectively~\citep{eichenbaum2007medial}. The most prominent instantiation of this dichotomy is the ``\textit{dual visual streams}" ~\citep{kravitz2011new, kravitz2013ventral}. Indeed, the ventral pathway projects through V4, IT, and perirhinal cortex into LEC, whereas the dorsal pathway projects through MT, PPC, and postrhinal/parahippocampal cortex into MEC (\textbf{Fig.~\ref{fig:overview}})~\citep{eichenbaum2007medial, kravitz2011new, kravitz2013ventral}. While traditionally the role of the ventral pathway has been extensively studied in terms of hierarchical object recognition, the integrative computation the dorsal pathway performs \textit{en route} to MEC, particularly along its high-order (parieto-medial) branch, remains poorly understood. This imbalance impedes a unified account of the dual visual streams, and leaves the origin of hippocampal generalization incompletely understood.

Notably, the root of this dichotomy can be traced back even to the retina level, where two functionally distinct ganglion-cell subtypes already establish the divergence: the Magnocellular (M) pathway conveys rapid signals of high temporal but low spatial resolution and feeds the dorsal stream, which specializes in spatiotemporal processing such as optical flow, depth, and saliency detection; conversely, the Parvocellular (P) pathway exhibits the opposite tradeoff (lower temporal but higher spatial resolution) and feeds the ventral stream, dedicated to recognizing sensory details. Given such extensive hierarchical depth of each stream,  we hereafter divide this pathway into two segments (\textbf{Fig.~\ref{fig:overview}}): the \textit{primary} pathway (the retina to low order cortical areas) and the \textit{extended} pathway (their continuation through higher-order associative cortices toward MEC/LEC).

Despite this anatomical continuity, so far the relational structures expressed in the entorhinal cortex—a core substrate for generalization, encoded by MEC grid cells—have largely been treated as a \textit{de-novo }product of local hippocampal-entorhinal computation, without larger-scale brain involvement \citep{dordek2016extracting, stachenfeld2017hippocampus, sorscher2023unified}.  Recent findings, however, increasingly refine this view: grid-like firing patterns are observed in the upstream of MEC, including the retrosplenial cortex and posterior cingulate cortex~\citep{doeller2010evidence,jacobs2013direct, constantinescu2016organizing}, both along the dorsal (parieto-medial) pathway. This suggests that the relational structure instantiated by grid codes may not arise solely within MEC, but be progressively shaped along the dorsal pathway, with potential roots in the functional dichotomy of the two visual streams. In other words, the two streams may \textit{co-emerge} as a complementary factorization of  visual processing, with the dorsal hierarchy progressively distilling relational structures from spatiotemporally organized M-pathway inputs, and the ventral hierarchy largely focusing on the representation of sensory-specific details from P-pathway inputs.

Here, we tested this hypothesis \emph{in silico} with \textit{Recurrent Graph} (\textbf{ReGraph}), a dual-stream model simulating two visual-processing stages (primary and extended; \textbf{Fig.~\ref{fig:architecture}}). The primary stage realizes the M/P-driven functional divergence through spatiotemporally asymmetric inputs, leveraging stream-specific pretrained networks: \textit{i)} depth, optical flow, and saliency estimation in the dorsal stream, and \textit{ii)} object-oriented processing in the ventral stream. The extended stage models their downstream pathways (routes to MEC/LEC) using architecturally homogeneous ReGraph layers that integrate representations via lateral and recurrent connectivities. We trained this model on the \textit{Something-Something V2} (SSV2) action-recognition dataset~\citep{goyal2017something}, a task neither pathway can solve alone, requiring context-invariant spatiotemporal regularities transferable to unseen action samples. To quantify whether relational structures indeed progressively crystallize along the extended stream during this task, we introduced \textbf{Relational Primitive Analysis} (\textbf{RPA}), designed to \textit{i)} evaluate the two essential primitives, context-invariance and gridness, and \textit{ii)} calculate both across the extended dorsal and ventral stages layer by layer. Finally, we probed the functional relationship of the two primitives, revealing how grid-like bases mechanistically support context-invariant generalization.

Our contributions are five-fold: \textit{i)} a dual-stream model jointly implementing the primary and extended visual pathways with parsimonious biological inductive biases, \textit{ii)} demonstration of top-down dorsal-to-ventral feedback effects; \textit{iii)} the RPA framework for quantifying layer-wise relational primitives; \textit{iv)} simulation evidence that such biological biases collectively support the progressive emergence of grid-like bases alongside context-invariant codes in the extended dorsal stream; and \textit{v)} ablation and reconstruction evidence that high-gridness bases in later dorsal layers support robust context-invariance and may serve as reusable routing templates for lateral connectivity. Together, these findings provide a mechanistic account suggesting that the computational basis for generalization may not be exclusively inherent in the hippocampal-entorhinal cortex, but rather emerge from the factorization of the dual visual streams.
\section{Methods}
\label{sec:method}
\begin{figure}[tbp]
    \centering
    \includegraphics[width=1\textwidth]{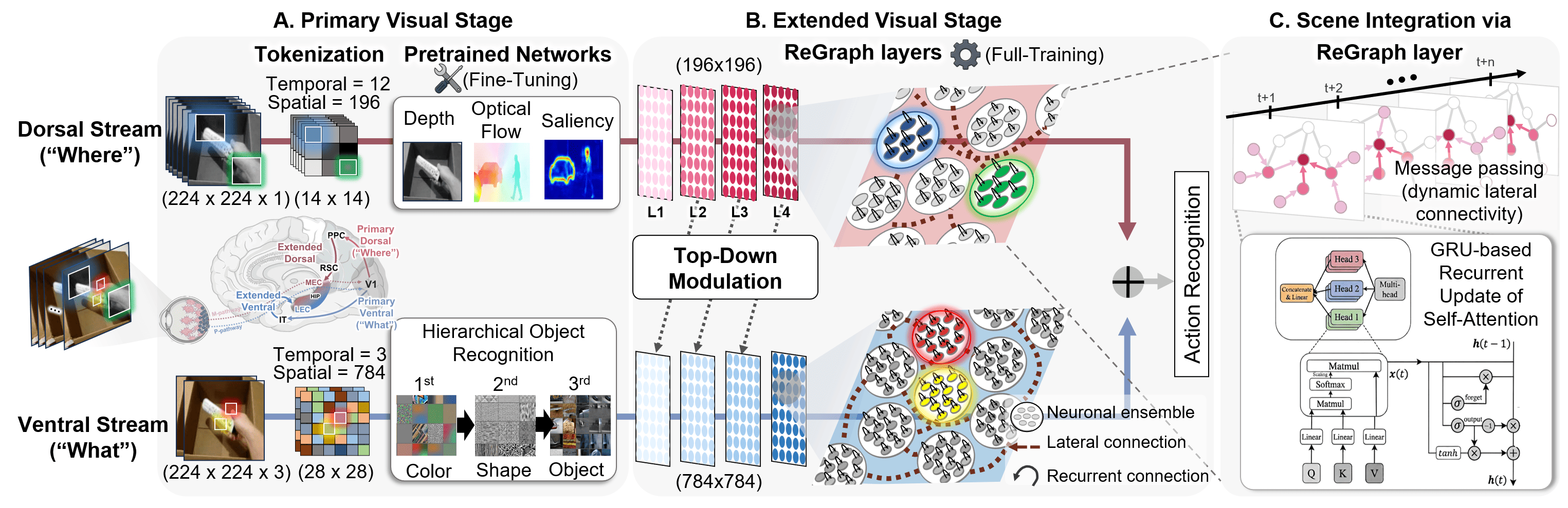}
    \caption{\textbf{The ReGraph architecture.} 
    \textbf{A.} Input is tokenized into M/P-driven asymmetric views (Dorsal: high temporal/low spatial; Ventral: low temporal/high spatial) and processed via specialized pretrained networks for spatiotemporal process (depth, optical flow, saliency) and hierarchical object-oriented processing.  
    \textbf{B.} Stacked layers model cortical hierarchies, with top-down modulation routing the dorsal signals to the ventral layers.  
    \textbf{C.} Self-attention acts as a dynamic graph operator, followed by a GRU-based recurrent update to regulate temporal flow.}
    \label{fig:architecture}
\end{figure}

We propose \textbf{ReGraph}, a dual-stream model that implements an integrated computational architecture of the dorsal and ventral pathways with biological details to study the emergence of relational primitives (\textbf{Fig.~\ref{fig:architecture}}). In each pathway, we built two sequential stages: First, the \textbf{Primary Visual Stage} (\S\ref{sec:primary}) simulates two essential biological details, \textit{i)} spatiotemporally asymmetric visual inputs due to different functional properties of retina's ganglion-cell subtypes (magnocellular [M] and parvocellular [P] cells), and \textit{ii)} the known computational principles of dorsal \textit{vs.} ventral visual pathways (spatiotemporal processing \textit{vs.} object-oriented processing, respectively). Second, the \textbf{Extended Visual Stage} (\S\ref{sec:extended}) models the higher-order cortical hierarchy using structurally homogeneous ReGraph layers that consist of lateral and recurrent connectivities. Finally, we also incorporated a \textbf{Dorsal-to-Ventral Modulation} mechanism (\S\ref{sec:topdown}), which integrates the dual streams by routing the signals from higher dorsal layers to lower ventral layers. 

\subsection{Primary visual stage: asymmetric input and stream-specific encoding}
\label{sec:primary}
\textbf{M/P-driven asymmetric input.} In the beginning, the sensory input enters the retina and is transmitted through the optical fibers to \textbf{V1}. In our model, this process starts with tokenization of each visual scene (\textit{e.g.,} video frame) into multiple spatial grid patches,  where each corresponds to a localized neuronal ensemble, consistent with topographic organization of V1 (retinotopy; \textbf{Fig.~\ref{fig:architecture}A}). Given the input frames $\mathbf{X} \in \mathbb{R}^{B \times T \times C \times H \times W}$ (\textit{B}=Batch size, \textit{T}=Temporal length, \textit{C}=Channel, \textit{H}=Height, \textit{W}=Width; for \textbf{SSV2}, $H=W=224$, see \S\ref{sec:RPA} below), we reshaped $\mathbf{X}$ into two functionally distinct views, reflecting the biological features of each M/P pathway: 

\begin{itemize}
    \item \textbf{Magnocellular pathway (M)}: Built on large cells with fast conduction, processes low spatial / high temporal resolution, achromatic luminance contrast (gray scale), and motion, projecting into the dorsal ("where") stream.
    \item \textbf{Parvocellular pathway} \textbf{(P)}: Built on small cells with slower conduction, processes high spatial / low temporal resolution, red–green chromatic information (RGB), and fine form and detail, projecting into the ventral ("what") stream.
\end{itemize}
To reflect these features, we parameterized the number of video frames per batch ($T$) and spatial tokens per frames ($N$) according to the M or P pathways: with a scaling factor $\alpha \geq 1$, we set $T_{Dorsal} = \alpha T_{Ventral}$ and $N_{Dorsal} = N_{Ventral}/\alpha$. In this way, the ventral stream uses fewer frames with denser RGB tokens, whereas the dorsal stream uses more frames with fewer grayscale tokens, approximating complementary M/P spatiotemporal profiles. In our model ($\alpha=4$), the dorsal stream samples $T=12$ frames, tokenizing each into a $14\times 14$ grayscale grid ($N=196$), whereas the ventral stream samples $T=3$ frames, tokenizing each into a $28\times28$ RGB grid ($N=784$).
 
\textbf{Stream-specific pretrained networks.} We continued to model the primary stage according to established functionalities of each visual pathway, that is, hierarchical object-oriented processing for the ventral stream, and spatiotemporal dynamics processing for the dorsal stream. For this, we employed existing pretrained models, each trained on one of these functionalities: The dorsal stream fuses intermediate embeddings from the networks pretrained on depth, optical flow, and saliency estimation, whereas the ventral stream takes those from an object-oriented processing network (\textbf{Fig.~\ref{fig:architecture}A}, \textit{Pretrained networks}; App.~\ref{app:extractors}). This yields a tokenized representation $\mathbf{E}_{\textit{Dorsal or Ventral}} \in \mathbb{R}^{B \times T \times N \times d}$ per stream, with shared embedding dimension $d=512$. 

\subsection{Extended visual stage: stream-generic ReGraph layers}
\label{sec:extended}
In the extended stage (\textit{i.e.,} higher-order visual pathways feeding the entorhinal cortex), we implemented a `\textit{stream-generic}' ReGraph network to further process and integrate $\mathbf{E}_{\textit{Dorsal or Ventral}}$. Importantly, the architecture of this network was built upon the four major biological principles detailed in the following subsections.
\begin{table}[t]
\centering
\caption{\textbf{Ablation over design choices.} All conditions share a fixed token budget ($T_D N_D + T_V N_V \approx 4{,}800$). Temporal ($T_D \!>\! T_V$) and spatial ($N_V \!>\! N_D$) imbalances realize the proposed dorsal and ventral traits, respectively. ID: \textit{In-distribution} evaluation.}
\label{tab:ablation}
\small
\renewcommand{\arraystretch}{1.15}
\setlength{\tabcolsep}{2pt}
\begin{tabular*}{\textwidth}{@{\extracolsep{\fill}}l l cc cc c c@{}}
\toprule
\textbf{Ablation Type} & \textbf{Variant} & $\mathbf{T_D / T_V}$ & $\mathbf{N_D / N_V}$ & \textbf{V-RGB} & $\mathbf{D{\to}V}$ & \textbf{Tokens} & $\substack{\textbf{Top-1} \\ \textbf{(\%; ID)}}$ \\
\midrule
\textit{Single-Stream}     & Dorsal-Only   & 12 / -- & 196 / --  & --  & --         & 2{,}352 & 71.69 \\
\textit{References}        & Ventral-Only  & -- / 3  & -- / 784  & Yes & --         & 2{,}352 & 60.38 \\
\midrule
\textit{Dorsal--Ventral}        & (a) Trait-Symmetric                        & 5 / 5           & 484 / 484          & No           & $\times$            & 4{,}840          & 69.78 \\
\textit{Biological Trait}   & (b) Dorsal-trait Only                   & 8 / 2           & 484 / 484          & No           & \checkmark          & 4{,}840          & 72.42 \\
\textit{Asymmetry}     & (c) Ventral-trait Only                  & 5 / 5           & 196 / 784          & Yes          & $\times$            & 4{,}900          & 70.43 \\
                           & (d) \textbf{Fully Asymmetric (ReGraph)} & \textbf{12 / 3} & \textbf{196 / 784} & \textbf{Yes} & \textbf{\checkmark} & \textbf{4{,}704} & \textbf{74.57} \\
\midrule
\textit{$D{\to}V$ Removal}  & Unmodulated   & 12 / 3  & 196 / 784 & Yes & $\times$   & 4{,}704 & 72.25 \\
\bottomrule
\end{tabular*}
\end{table}

\textbf{1) Intra-areal lateral, input-driven dynamic connectivity.} We implemented this first principle---\textit{input-driven, dynamically changing intra-areal lateral connectivity}---using a Graph Neural Network (GNN) paradigm, which expresses spatially distributed neuronal ensembles as node ($v_i$) and their links via edge weights $\mathbf{e}_{ji}$, summarized globally by the adjacency matrix ($\hat{\mathbf{A}}$). Among GNN variants, \textit{Graph Convolutional Networks} (\textit{GCNs})~\citep{kipf2016semi} compress GNN message passing into a single matrix product:
\begin{equation}
\mathbf{H}^{\ell+1}
\;=\;
\sigma\!\big(\,
    \underbrace{\hat{\mathbf{A}}}_{\substack{\text{adjacency} \\ \text{matrix}}}
    \underbrace{\mathbf{H}^{\ell}\,\mathbf{W}}_{\substack{\text{node} \\ \text{signal}}}
\,\big)
\label{eq:gcn}
\end{equation}
where the (normalized) adjacency matrix $\hat{\mathbf{A}}$ dictates how strongly each node's signal flow to every other node; $\mathbf{H}^{\ell}\mathbf{W}$ is the node signal traveling along this network, obtained by a learnable linear transformation ($\mathbf{W}$) of the current state; $\sigma$ is the non-linear activation function. In sum, the matrix product $\hat{\mathbf{A}}\,\mathbf{H}^{\ell}\mathbf{W}$ performs lateral aggregation for every node in parallel, making GCN an interpretable, biologically suggestive operator. 
The only problem in the standard GCN model is that $\hat{\mathbf{A}}$ is predefined, sparse, and static---none of which is consistent with the local neural circuits. In fact, any two neuronal ensembles can in principle influence one another given the right context or external perturbation (\textit{e.g.,} sensory input). Restricting interactions to a fixed local neighborhood thus discards the very phenomenon we aim to model. Addressing this requires the adjacency to be \textit{i)} defined over all node pairs (\emph{dense and global}) and \textit{ii)} recomputed from the current state at every forward pass (\emph{input-driven dynamic}). Notably, Multi-Head Self-Attention (MHSA)~\citep{vaswani2017attention} naturally fulfills all these requirements as a generalization of the GCN operator.

\begin{equation}
    \sigma\Big( 
    \underbrace{\vphantom{\mathbf{H}^{\ell}\mathbf{W}} \hat{\mathbf{A}}}_{\substack{\text{static} \\ \text{adjacency} \\ \text{matrix}}} 
    \underbrace{\mathbf{H}^{\ell}\mathbf{W}}_{\substack{\text{node} \\ \text{signal}}} 
    \Big) 
    \;\;\Longleftrightarrow\;\; 
    \underbrace{\mathrm{softmax}\Big(\frac{\mathbf{Q}\mathbf{K}^{\top}}{\sqrt{d_k}}\Big)}_{\substack{\text{dynamic} \\ \text{adjacency matrix}}} 
    \underbrace{\vphantom{\mathrm{softmax}\Big(\frac{\mathbf{Q}\mathbf{K}^{\top}}{\sqrt{d_k}}\Big)} \mathbf{V}}_{\substack{\text{node} \\ \text{signal}}},
\quad
\mathbf{Q}\!=\!\mathbf{H}^{\ell}\mathbf{W}_Q,\;
\mathbf{K}\!=\!\mathbf{H}^{\ell}\mathbf{W}_K,\;
\mathbf{V}\!=\!\mathbf{H}^{\ell}\mathbf{W}_V.
\label{eq:gcn_to_attn}
\end{equation}

As shown in Eq.~\ref{eq:gcn_to_attn}, MHSA replaces GCN's static $\hat{\mathbf{A}}$ with the row-wise softmax of $\mathbf{Q}\mathbf{K}^{\top}$, yielding a \emph{dynamic adjacency matrix} that is dense over all $N\times N$ node pairs and computed from the current input $\mathbf{H}^{\ell}\!\leftarrow\!\mathbf{E}_{\textit{Dorsal or Ventral}}$ (\S\ref{sec:primary}). These inputs are shaped by the learnable projections $\mathbf{W}_Q$ and $\mathbf{W}_K$, which determine \emph{what counts as a connection in the first place}. The softmax provides row-wise normalization analogous to the degree-based normalization in $\hat{\mathbf{A}}$. The value projection $\mathbf{V}=\mathbf{H}^{\ell}\mathbf{W}_V$ plays the role of $\mathbf{H}^{\ell}\mathbf{W}$ in GCN, carrying the lateral signal along the dynamically inferred connectivity. The model thus recomputes its message-passing graph each time new sensory patches arrive, realizing input-driven dynamic connectivity.

\textbf{2) Multi-channel parallel processing.} A further advantage of MHSA emerges from its multi-head structure: in our study, with $H=8$ heads operating in parallel, a single ReGraph layer instantiates several distinct lateral routing patterns simultaneously. Biologically, this mirrors a hallmark of cortical computation---the recruitment of multiple, partially redundant populations processing the same input in parallel for robustness and richer feature decomposition. This operator yields the laterally aggregated features $\tilde{\mathbf{E}}^{\ell}_{h}$ for each head $h$ and layer $\ell$. Together, principles \textbf{(1)} and \textbf{(2)} argue that MHSA is not an ad-hoc import but the \emph{minimal} modification of graph message-passing that realizes two cortical strategies---dynamically changing information routing and multi-channel parallel processing. A component-by-component reformulation of GCN and Self-Attention within the conventional message-passing framework is provided in App.~\ref{app:sec:graph_framework}.

\textbf{3) Recurrent connections and cortical hierarchy.} To model the third principle---\textit{temporal retention via recurrent connections}---we pass $\tilde{\mathbf{E}}^{\ell}_{h}$ through a per-head Gated Recurrent Unit (GRU): the recurrence supports temporal integration, while multiplicative gates (${z_t, r_t}$) act as a computational proxy for biological gating, selectively retaining or overwriting prior state at each step. This yields the temporally updated representations $\hat{\mathbf{E}}^{\ell}_{h}$ (App.~\ref{app:sec:gated_recurrence}). Finally, the concatenated output from all heads undergoes additional intra-layer operations (\textit{e.g.,} normalization, dropout, residuals) and is propagated forward via a learned linear projection mapping $N_\ell$ tokens to $N_{\ell+1}$ tokens through inter-areal hierarchical connections.

\begin{table}[!t]
\centering
\caption{\textbf{OOD linear probing for context-invariance.} \textbf{OOD [Drop]}: OOD accuracy (\%) and its train--test gap (pp), averaged over 8 heads (30 samples/class). Bold: superior OOD accuracy in the dorsal streams (avg. $>$80\%) and the L3--L4 surge in the ventral stream of \emph{ReGraph} ($D{\to}V$).}
\label{tab:linear_probe}
\renewcommand{\arraystretch}{1.1}
\resizebox{\textwidth}{!}{%
\begin{tabular}{ll ccccc}
\toprule
\textbf{Model} & \textbf{Stream} & \textbf{L1} (OOD [Drop]) & \textbf{L2} (OOD [Drop]) & \textbf{L3} (OOD [Drop]) & \textbf{L4} (OOD [Drop]) & \textbf{Avg} (OOD [Drop]) \\
\midrule
Dorsal-Only & Dorsal & 72.8$\pm$3.6 [$-$25.6] & 69.0$\pm$3.7 [$-$30.7] & 76.7$\pm$1.9 [$-$23.2] & 93.6$\pm$1.6 [$-$6.4] & 78.0 [$-$21.5] \\
\midrule
\multirow{2}{*}{Unmodulated}
  & Dorsal  & \textbf{86.8$\pm$2.7 [$-$12.7]} & \textbf{82.9$\pm$2.1 [$-$17.1]} & \textbf{81.5$\pm$2.1 [$-$18.5]} & \textbf{96.9$\pm$0.4 [$-$3.1]} & \textbf{87.0 [$-$12.9]} \\
  & Ventral & 52.0$\pm$3.8 [$-$44.2] & 56.9$\pm$2.7 [$-$42.1] & 59.2$\pm$4.5 [$-$40.2] & 65.4$\pm$1.7 [$-$33.4] & 58.4 [$-$40.0] \\
\midrule
\multirow{2}{*}{\shortstack[l]{ReGraph}}
  & Dorsal  & \textbf{82.0$\pm$2.7 [$-$16.9]} & \textbf{81.5$\pm$3.2 [$-$18.5]} & \textbf{78.5$\pm$3.4 [$-$21.3]} & \textbf{97.4$\pm$0.6 [$-$2.6]} & \textbf{84.9 [$-$14.8]} \\
  & Ventral & 55.0$\pm$5.7 [$-$43.6] & 59.7$\pm$1.7 [$-$40.0] & \textbf{95.9$\pm$1.1 [$-$4.0]} & \textbf{95.0$\pm$2.5 [$-$4.9]} & 76.4 [$-$23.1] \\
\bottomrule
\end{tabular}%
}
\end{table}

\textbf{4) Top-down modulation: the Magnocellular Advantage.} 
\label{sec:topdown}
To implement the fourth principle, namely the \textit{Magnocellular Advantage} hypothesis~\citep{laycock2007role}---where rapid dorsal signals project forward to shape slower ventral processing---we introduced a cross-attention top-down modulation step in the ventral stream prior to its inter-areal propagation. The ventral MHSA output $\tilde{\mathbf{E}}^{\ell}_{h}$ acts as queries, while the dorsal representations from the higher layer $\ell+1$ provide keys and values. This enables the ventral stream to retrieve the most relevant higher-level dorsal context $\tilde{\mathbf{C}}^{\ell}_{h}$ (App.~\ref{app:sec:alignment}). The modulated ventral representation is then computed as:
\begin{equation}
    \tilde{\mathbf{E}}'^{\,\ell}_{h} = \tilde{\mathbf{E}}^{\ell}_{h} + \sigma(\gamma_{\ell,h})\tilde{\mathbf{C}}^{\ell}_{h}
    \label{eq:gated_addition}
\end{equation}

where $\sigma(\gamma_{\ell,h})$ is a learnable per-head gate initialized small ($\approx 0.25$) to preserve fine ventral details early in training. The modulated $\tilde{\mathbf{E}}'^{\,\ell}_{h}$ then proceeds through the recurrent and inter-layer stages up the hierarchy. After $L$ layers, terminal representations from both streams are spatiotemporally pooled and combined for downstream classification (App.~\ref{app:classification}).

\subsection{RPA pipeline: quantifying relational primitives}
\label{sec:RPA}

SSV2~\citep{goyal2017something} is a benchmark for testing relational primitives: each class is defined by a templated hand--object interaction (e.g., \emph{"pushing [something] from left to right"}), and class identity depends on the spatiotemporal hand--object relation (App. Fig.~\ref{fig:ssv2_examples}). Within-class instances vary in object instantiation, appearance, viewpoint, and speed, requiring models to abstract beyond superficial visual cues. Correct classification thus requires extracting a context-invariant hand--object relation generalizing across these variations---the relational structure central to our analysis. With the dual-stream model trained on this task, we sought to quantify the progressive emergence of relational primitives along each visual pathway, especially in the extended stage. To this end, we developed the Relational Primitive Analysis (RPA) pipeline, which operationally defines the relational primitive as a representation jointly satisfying two criteria: first, the dominant pattern of information flow shaped by dynamic intra-areal lateral connectivity at each layer exhibits a grid-like topology (\emph{gridness}); second, the hidden representations updated through this connectivity generalize to unseen samples (\emph{context-invariance}). Building upon these definitions, the pipeline finally probed the functional relationship between these two primitives to understand their computational mechanism, with details described below. 

\textbf{1) Gridness in the bases of lateral connectivity.}
\label{sec:grid}
(\textbf{App.~Fig.~\ref{fig:RPA}A}). Adapting the analytical framework from \citep{stachenfeld2017hippocampus}, we treated the attention matrix (before softmax; `pre-softmax') for each layer as a `topographic' adjacency matrix, where nodes represent spatially anchored tokens and edges denote their connectivity strengths. Crucially, our target was not the connectivity matrix itself but its dominant bases---the principal spatial templates along which lateral interactions are organized. In the entorhinal cortex, grid cells emerge precisely as such dominant eigenmodes of place-to-place transitions~\citep{stachenfeld2017hippocampus}; here, extracting analogous bases of ReGraph's attention matrix therefore lets us ask whether comparable spatial primitives arise from its learned lateral connectivity. Given this motivation, we extracted the bases of our dynamic adjacency matrix (\S\ref{sec:extended}, principle 1) using Higher-Order Singular Value Decomposition (HOSVD)~\citep{de2000multilinear}. After retaining the top $3, 5, 7\%$ of these bases by their singular values, we reshaped them into 2D layouts to yield spatial weighting maps (analogous to grid cell firing-rate maps in \citep{stachenfeld2017hippocampus}; see \textbf{App.~\ref{sec:bg_grid}}). We then computed the gridness score~\citep{sargolini2006conjunctive} for each map. A basis exhibits a valid grid-like pattern (a \emph{grid-like spatial basis}) if its score exceeds a significance threshold derived from $100$ spatial shuffles per map, following~\citet{banino2018vector}. We reported the global ratio of grid-like spatial bases within the retained pool per layer and stream (\textbf{App.~\ref{app:gridness}}).

\textbf{2) Context-invariance of hidden representations.} 
\label{sec:ood}
(\textbf{App.~Fig.~\ref{fig:RPA}B}). While the grid analysis examines the structural topology of lateral connectivity ($\mathbf{Q}\mathbf{K}^{\top}$), we further evaluated whether the ReGraph's intermediate representations generalizes to unseen cases. To this end, we pooled updated representations across space and time, yielding a single summary feature vector per sample, layer, and attention head. Following the standard linear probing protocol~\citep{alain2016understanding}, we trained a logistic-regression classifier on these vectors to predict action classes under an out-of-distribution (OOD) split where train and test sets share no object identities. Per-head accuracies are averaged into a layer-wise score per stream, enabling cross-stream comparison. The train-to-OOD accuracy gap serves as an inverse proxy for context-invariance: smaller gaps indicate higher generalization driven by context-invariant spatiotemporal structure rather than object-specific cues.

\textbf{3) Functional relevance of grid-like bases.}
\label{sec:causal}
(\textbf{App.~Fig.~\ref{fig:relevance}}).
The aforementioned analyses establish how to independently measure gridness and context-invariance. Building on this, we performed two additional analyses to test whether the grid-like bases genuinely support context-invariant generalization, and to uncover the mechanism by which they carry this information. First, isolating the top $5\%$ of attention bases by their singular values as in \textbf{1)} above, we ablated bases with high ($G_{\mathrm{high}}$) and low ($G_{\mathrm{low}}$) gridness scores (\emph{\textbf{i. Bases Ablation}}\label{method:ablation}). We then measured the resulting shifts in OOD accuracy, hypothesizing that the ablation of bases carrying more context-invariant information would lead to a larger performance drop. To quantify which basis set induces a more severe degradation, we define the accuracy gap between the two ablation settings as $\Delta_{Acc.} = \text{Acc.}_{G_{\mathrm{low}}} - \text{Acc.}_{G_{\mathrm{high}}}$. We also tested whether these results remain consistent under varying levels of noise perturbation to ensure they are not merely coincidental. Second, to uncover the computational mechanism, we tested which basis sets can better reconstruct unseen attention matrices (\emph{\textbf{ii. Connectivity Reconstruction}}\label{method:recon}). To quantify this, we defined the reconstruction gap as $\Delta_{R^2} = R^2_{G_{\mathrm{high}}} - R^2_{G_{\mathrm{low}}}$, where $R^2$ denotes our defined reconstruction score. Crucially, for both metrics, a larger positive gap ($\Delta_{Acc.}$ and $\Delta_{R^2}$) indicates that the model relies more heavily on $G_{\mathrm{high}}$ than $G_{\mathrm{low}}$—both for maintaining context-invariant generalization and for capturing the underlying connectivity patterns. The results are summarized in Table \ref{tab:relevance}. Detailed protocols for both analyses are provided in \textbf{App. ~\ref{relevance}}.
\section{Experiments and results}
\label{sec:experiments}
\subsection{ReGraph validation and design-choice ablation}
\label{subsec:setup_ablation}

\textbf{ReGraph validation.} Our dual-stream model comprises two stages (\textbf{Fig.~\ref{fig:architecture}}): a \emph{primary visual stage}, initialized from pretrained M-/P-cell pathways and task-specific backbones, and an \emph{extended visual stage} of fully trainable \emph{ReGraph} layers. The principal training target was the extended stage: all \emph{ReGraph} layers are trained from scratch at a learning rate (LR) of $3 \times 10^{-4}$. Pretrained backbones in the primary stage are also fine-tuned, but mildly---at a $10\times$ smaller LR ($3 \times 10^{-5}$)---to adapt to SSV2 while preserving overall feature spaces shaped by original pretraining. This reduced LR is a standard method in the field, used when jointly optimizing pretrained and fully trainable modules~\citep{carion2020end}. We evaluated \textbf{ReGraph} on a 33-class SSV2 subset (\textbf{App.~\ref{app:training}}) under an in-distribution setting. Before diving into the RPA, we validated the quality of our learned representations against established baseline algorithms for an action recognition task: SlowFast~\citep{feichtenhofer2019slowfast}, TimeSformer~\citep{bertasius2021space}, and VideoMAE~\citep{tong2022videomae}. In this evaluation, ReGraph yielded a top-1 accuracy of $74.57\%$. While VideoMAE set a higher upper bound ($80.4\%$), ReGraph took a strong intermediate position by exceeding both SlowFast ($66.1\%$) and TimeSformer ($62.3\%$). This performance comparison suggests that the learned feature space is robust enough for subsequent RPA (see \textbf{App.~\ref{app:baseline_comparison}} for further details).

\begin{table}[t]
\centering
\caption{\textbf{Ratio of grid-like bases across layers.} We report the fraction of top 5\% singular-value bases exceeding a shuffled null threshold (\%), formatted as \textbf{Dorsal / Ventral} (App.~\ref{app:gridness}). Bold indicates a strictly increasing trend unique to \emph{ReGraph}'s extended dorsal stream.}
\label{tab:gridness_main}
\footnotesize
\setlength{\tabcolsep}{4pt}
\begin{tabular*}{\textwidth}{@{\extracolsep{\fill}} l ccccc @{}}
\toprule
\textbf{Model} & \textbf{L1} & \textbf{L2} & \textbf{L3} & \textbf{L4} & \textbf{Avg} \\
\midrule
Dorsal-Only & 0.0 / -- & 5.6 / -- & 4.0 / -- & 2.6 / -- & 3.0 / -- \\
Unmodulated & 0.1 / 0.1 & 8.6 / 6.8 & 4.0 / 5.9 & 6.3 / 5.9 & 4.7 / 4.7 \\
ReGraph     & \textbf{0.0} / 0.1 & \textbf{5.1} / 6.5 & \textbf{8.8} / 6.0 & \textbf{15.3} / 7.2 & \textbf{7.3} / 4.9 \\
\bottomrule
\end{tabular*}
\end{table}

\begin{figure}[t]
    \centering
    \includegraphics[width=\textwidth, trim=8 246 8 40, clip]{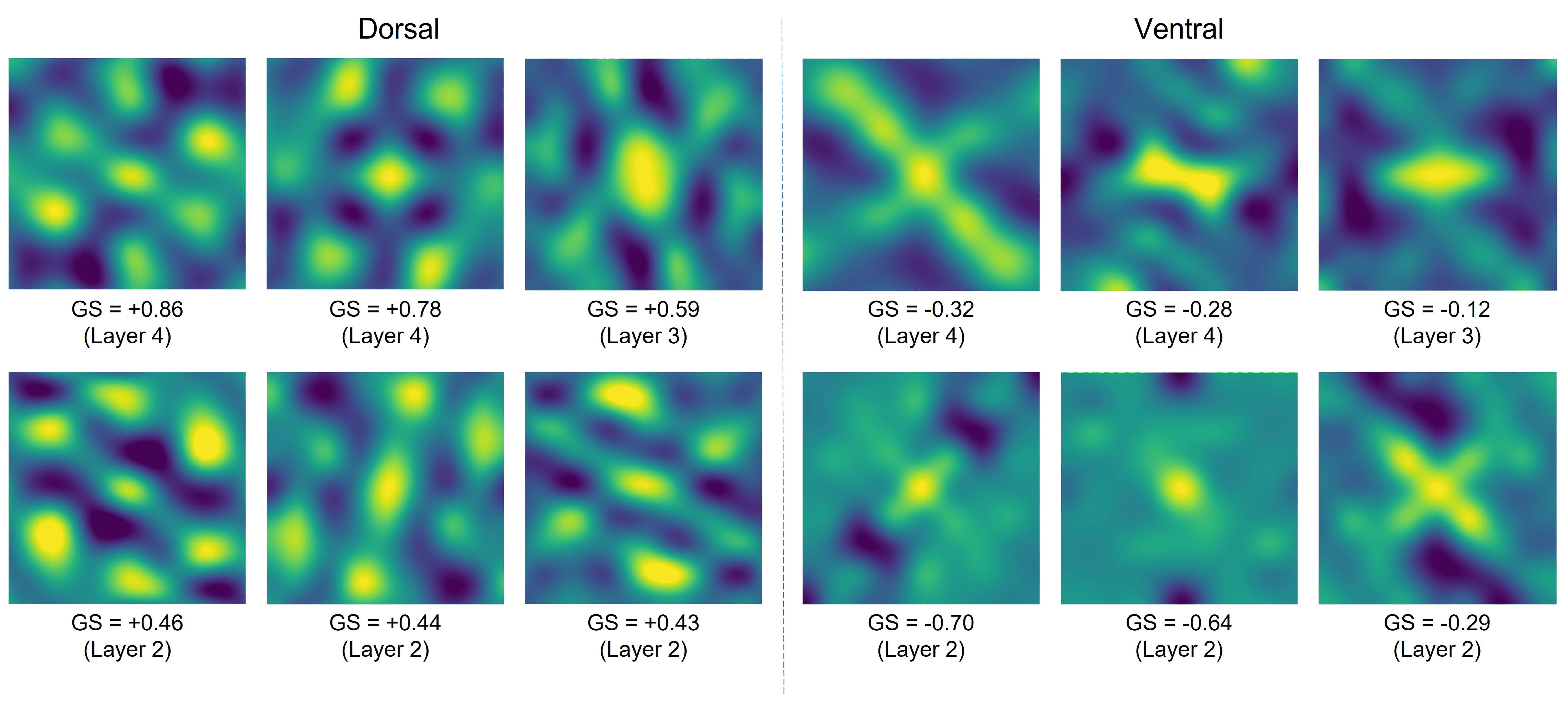}
    \caption{\textbf{Spatial weighting maps in \emph{ReGraph}.} 2D autocorrelograms of pre-softmax attention bases; GS: gridness score. \textbf{Dorsal (left three)}: hexagonal patterns with varied orientation and spacing. \textbf{Ventral (right three)}: irregular or cross-like patterns lacking hexagonal structure, with near-zero GS.}
    \label{fig:gridness_vis}
\end{figure}

\textbf{Design-choice ablation.} We next examined how each architectural choice contributes to our dual-stream model performance  via a controlled ablation across three categories (Table~\ref{tab:ablation}): \textbf{\textit{1) Single-Stream References}}: We compared two single-stream references—\emph{Dorsal-Only} and \emph{Ventral-Only}—retaining only the corresponding pathway. \textbf{\textit{2) Dorsal--Ventral Biological Trait Asymmetry}}: We also evaluated configurations selectively activating the asymmetric biological trait of one pathway, both, or neither in the dual-stream architecture. In other words, the \emph{Dorsal-trait Only} variant preserves only faster temporal sampling and $D{\to}V$ modulation (no ventral traits), whereas the \emph{Ventral-trait Only} keeps only higher spatial resolution with RGB encoding (no dorsal traits). This yields four variants: \emph{Symmetric} (neither trait), \emph{Dorsal-Trait Only}, \emph{Ventral-Trait Only}, and ReGraph (both traits, fully asymmetric). \textbf{\textit{3) D$\rightarrow$V (Modulation) Removal}:} We further isolated the contribution of cross-stream feedback by disabling $D{\to}V$  modulation in the full ReGraph (\emph{Unmodulated}). For `Single-Stream References' category, \emph{Dorsal-Only} ($71.69\%$) and \emph{Ventral-Only} ($60.38\%$) showed reduced performance compared to the full model jointly processing both dorsal and ventral traits (\textbf{ReGraph}) ($74.57\%$), confirming the benefit of stream coupling (Table~\ref{tab:ablation}). For `Dorsal--Ventral Biological Trait Asymmetry' test, disabling both traits (\emph{Trait-Symmetric}) reduced performance to $69.78\%$, while activating either alone yielded partial improvements (\emph{Ventral-trait Only}: $70.43\%$; \emph{Dorsal-trait Only}: $72.42\%$). Eliminating D$\rightarrow$V top-down modulation reduced performance to $72.25\%$, highlighting the contribution of this cross-stream pathway in ReGraph.

\subsection{Layer-wise emergence of relational primitives via RPA}
\label{subsec:rpa}

In the previous section, since \emph{Dorsal-Only} and \emph{Unmodulated} variants emerged as the strongest alternatives, we included these two models alongside the proposed \textbf{ReGraph} for subsequent layer-wise RPA of relational primitive emergence along the extended streams (\textbf{App.~Fig.~\ref{fig:RPA}-~\ref{fig:relevance}}).

\textbf{1) Context-invariance via OOD linear probing.} We examined how representations evolve across dorsal ($D_{k}$) and ventral ($V_{k}$) layers, where $k$ indexes layer depth within each stream. A clear pattern emerged: in both the \textit{Unmodulated} and \textit{ReGraph} models, the dorsal streams exhibited strong context-invariance, showing substantially smaller average drops ($-12.9$ and $-14.8$ percentage points, pp) than ventral counterparts ($-40.0$ and $-23.1$ pp; \textbf{Table~\ref{tab:linear_probe}}, `Avg' column). Notably, however, this ventral stream in ReGraph exhibited a distinct layer-wise trajectory. While its early layers ($V_1$--$V_2$) indeed showed large drops ($\approx -40$ pp), the gap sharply diminished in the later layers, reaching $-4.0$ and $-4.9$ pp at ($V_3$--$V_4$), with OOD accuracies of $95.9\%$ and $95.0\%$. This transition coincided with the progressive opening of the $D{\to}V$ gate $\sigma(\gamma)$ (Eq.~\ref{eq:gated_addition}, a mechanism that regulates how much dorsal information is injected into the ventral stream), which increased from near initialization ($\approx 0.28$) to $0.47 \pm 0.02$ at $V_3$ (App. Table~\ref{tab:gate_values}). This gating dynamic reveals the underlying mechanism: context-invariance does not emerge independently in both pathways. Instead, it is computed natively within the dorsal stream, and the ventral stream acquires this property only at later stages by explicitly importing dorsal representations via cross-stream modulation.

\begin{table}[htbp]
\centering
\caption{\textbf{Functional relevance of grid-like bases.} Each cell reports results obtained under $G_{\text{high}}$ / $G_{\text{low}}$ manipulations [gap $\Delta_{Acc. / R^2}\uparrow$]. A larger $\Delta$ indicates greater reliance on $G_{\text{high}}$ for both tasks (OOD Acc. under ablation; reconstruction score). \textbf{Bold} ($\Delta > \text{row mean}$) highlights the L3-4 surge.}
\label{tab:relevance}
\resizebox{\columnwidth}{!}{%
\begin{tabular}{llcccc}
\toprule
\multicolumn{2}{l}{\textbf{Task}} & \textbf{L1 ($G_{\text{high}} / G_{\text{low}} [\Delta\uparrow]$)} & \textbf{L2 ($G_{\text{high}} / G_{\text{low}} [\Delta\uparrow]$)} & \textbf{L3 ($G_{\text{high}} / G_{\text{low}} [\Delta\uparrow]$)} & \textbf{L4 ($G_{\text{high}} / G_{\text{low}} [\Delta\uparrow]$)} \\
\midrule
\multirow{4}{*}{\shortstack[l]{\textbf{Bases}\\\textbf{Ablation}}} 
 & $\sigma = 0$ & 82.0 / 81.7 [$-0.3$] & 82.6 / 82.9 [$+0.3$] & 70.3 / 77.3 [\textbf{+7.0}]  & 89.1 / 95.1 [\textbf{+6.0}]  \\
 & $\sigma = 0.5$            & 70.3 / 70.1 [$-0.2$] & 75.3 / 75.4 [$+0.2$] & 61.7 / 70.2 [\textbf{+8.5}]  & 85.8 / 93.6 [\textbf{+7.7}]  \\
 & $\sigma = 1.0$            & 54.3 / 54.4 [$+0.1$] & 60.9 / 60.2 [$-0.7$] & 47.1 / 55.8 [\textbf{+8.7}]  & 75.4 / 86.9 [\textbf{+11.5}] \\
 & $\sigma = 1.5$            & 42.0 / 42.0 [$0.0$]  & 47.5 / 46.1 [$-1.4$] & 35.9 / 43.4 [\textbf{+7.5}]  & 61.2 / 75.2 [\textbf{+14.0}] \\
\midrule[\heavyrulewidth]
\multicolumn{2}{l}{\textbf{Reconstruction ($R^2$)}} & .000 / .014 [$-.014$] & .008 / .001 [$+.007$] & .082 / .000 [\textbf{+.082}] & .131 / .037 [\textbf{+.094}] \\
\bottomrule
\end{tabular}%
}
\end{table}

\textbf{2) Emergence of grid-like spatial weight patterns.} We next applied HOSVD (\S\ref{sec:grid}) to the dynamically changing attention matrices of our dual-stream model. For robustness, matrices were extracted only when processing the action classes with above-median accuracy. Using 30 correctly classified training samples per class, we analyzed each head and layer separately. For each class--head pair, the resulting bases were sorted by singular value magnitude, retaining the top 5\% of the total number of bases. We then aggregated significant grid-like base counts across classes and heads within each layer, yielding their proportion in this controlled subset (robustness checks at 3\% and 7\% retention rates in App. Table~\ref{tab:gridness_robustness}). This analysis showed a clear divergence across variants: Only the dorsal stream in \emph{ReGraph} showed a consistent monotonic increase in grid-like bases across layers ($0.0\% \rightarrow 5.1\% \rightarrow 8.8\% \rightarrow 15.3\%$; Table~\ref{tab:gridness_main}), whereas the \emph{Unmodulated} and \emph{Dorsal-Only} variants showed no such trend. Importantly, the aforementioned standard baselines (VideoMAE, TimeSformer, SlowFast) also failed to exhibit such a trend despite their competitive performance (App. Table~\ref{tab:baseline_gridness}). This contrast was finally corroborated in a base firing pattern of the models as well (\textbf{Fig.~\ref{fig:gridness_vis}}): dorsal bases frequently formed grid-like patterns—a central peak with six neighbors at regular angles—whereas ventral bases showed weaker or irregular responses lacking clear hexagonal organization.

\textbf{3) Functional role of grid-like bases:} Finally, we investigated whether grid-like bases act as the genuine computational substrate for context-invariant generalization. Our \emph{\hyperref[method:ablation]{Bases-Ablation}} analysis revealed that while the OOD accuracy difference ($\Delta_{\mathrm{Acc.}}$) between $G_{\mathrm{high}}$ and $G_{\mathrm{low}}$ is negligible in early layers (Table~\ref{tab:relevance}), late layers critically rely on $G_{\mathrm{high}}$. This reliance becomes even more pronounced under severe input noise, where $\Delta_{\mathrm{Acc.}}$ surges to +7.5 at L3 and +14.0 at L4 ($\sigma=1.5$), highlighting the inherent noise robustness of $G_{\mathrm{high}}$-driven representations. Delving into the mechanism, our \emph{\hyperref[method:recon]{Connectivity-Reconstruction}} analysis (\S\ref{sec:causal}) demonstrated a matching trend, with $\Delta_{R^2}$ increasing significantly towards L4. This indicates that these bases function as essential templates to reconstruct dynamically changing lateral connectivity, thereby facilitating generalization. Intriguingly, this escalating functional reliance perfectly mirrors the growing proportion of grid-like bases across layers (Table~\ref{tab:gridness_main}). Such synchronized emergence implies that the layer-wise accumulation of these robust, invariant templates ultimately drives the peak OOD accuracy observed at L4 (Table~\ref{tab:linear_probe}).

Together, our RPA revealed that, among the tested variants, only the full \emph{ReGraph} combines an increasing trend in grid-like bases with strong context-invariance along its extended dorsal stream. The lack of such a trend in \emph{Dorsal-Only} highlights the contribution of the primary ventral pathway's processing of sensory details, while its absence in \emph{Unmodulated} underscores the importance of cross-stream modulation beyond the mere coexistence of the two streams. Complementary comparisons with standard baselines, which show no comparable trend across layers (App.~\ref{app:baseline_comparison}), further support the hypothesis that our primary- and extended-stage biological inductive biases jointly provide a computational basis for the progressive emergence of grid-like patterns. Finally, our ablation and reconstruction analyses provide a mechanistic account that high-gridness bases indeed functionally support robust context-invariance in the later dorsal layers.
\section{Conclusion and limitations}
\label{sec:concolusion}
In this work, we assessed a potential hypothesis of how the relational structure could emerge in the medial temporal lobe, extending beyond the prior functional mapping approaches. Through ReGraph---a dual-visual-stream architecture grounded in neurobiological principles---and the RPA pipeline, we showed that major relational primitives progressively crystallize along the extended dorsal stream. This finding suggests that relational structure may not arise \emph{de novo} within the entorhinal-hippocampal system, but may instead be rooted in the early sensory divergence of the M and P pathways. Our framework has two main limitations that motivate future research. First, ReGraph operationalizes the Magnocellular Advantage via unidirectional modulation ($D \to V$); since biological cortices feature reciprocal connectivity, future work should integrate bidirectional communication ($D \leftrightarrow V$) to fully capture inter-stream dynamics. Second, our current evaluation focuses on relational structures derived from spatiotemporal visual stimuli; since generalization and relational structure fundamentally extend to abstract domains, assessing whether this dual-stream paradigm extends to higher-level non-visual reasoning tasks is a critical direction for future work.

\subsection*{AI use statement}

In this work, we did not use generative AI tools for any tasks requiring
disclosure, including the development of theoretical models or conceptual
frameworks, the formulation of hypotheses or mathematical claims, the design
of research methodology or experiments, the implementation of methods, data
generation or processing, or the interpretation of results. Generative AI
tools were used solely to edit the manuscript for grammar, clarity, and
readability. All AI-assisted edits were reviewed by the authors to verify
that the original meaning, claims, and technical content were preserved. We take responsibility for the final content of this work, including text, claims or artifacts produced with the aid of generative AI.

\subsection*{Ethics statement}

This work is foundational research aimed at understanding how relational
structures underlying generalization emerge in the brain, using publicly
available benchmark data (Something-Something V2) without involving human
subjects or personally identifiable information. We nevertheless acknowledge
a potential dual-use concern: the ability to recognize actions invariantly
to background context or object identity could, in principle, be repurposed
for surveillance or invasive monitoring applications. To mitigate this risk,
our public release is limited to research-purpose code and checkpoints
trained on a constrained 33-class hand--object interaction subset of SSV2,
rather than a deployment-ready recognition pipeline. A more detailed discussion is provided in App.~\ref{app:broader_impact}.

\subsection*{Reproducibility statement}

Our source code is available at
\url{https://anonymous.4open.science/r/regraph-351D/README.md}, and pretrained
model weights at \url{https://doi.org/10.5281/zenodo.19917046}. Full
training details and hyperparameters are provided in
App.~\ref{app:training}, the architectures and selection criteria of
the stream-specific pretrained feature extractors in
App.~\ref{app:extractors}, the complete procedure of the RPA gridness
analysis in App.~\ref{app:gridness}, the baseline training and analysis
protocol in App.~\ref{app:baseline_comparison}, the bases-ablation and
connectivity-reconstruction protocols in App.~\ref{relevance}, and the
computational resources
used in App.~\ref{app:compute}.

\bibliography{iclr2027_conference}
\bibliographystyle{iclr2027_conference}

\appendix
\section*{Appendix}

\section*{Content Overview}
\begin{figure} 
    \centering
    \includegraphics[width=\textwidth]{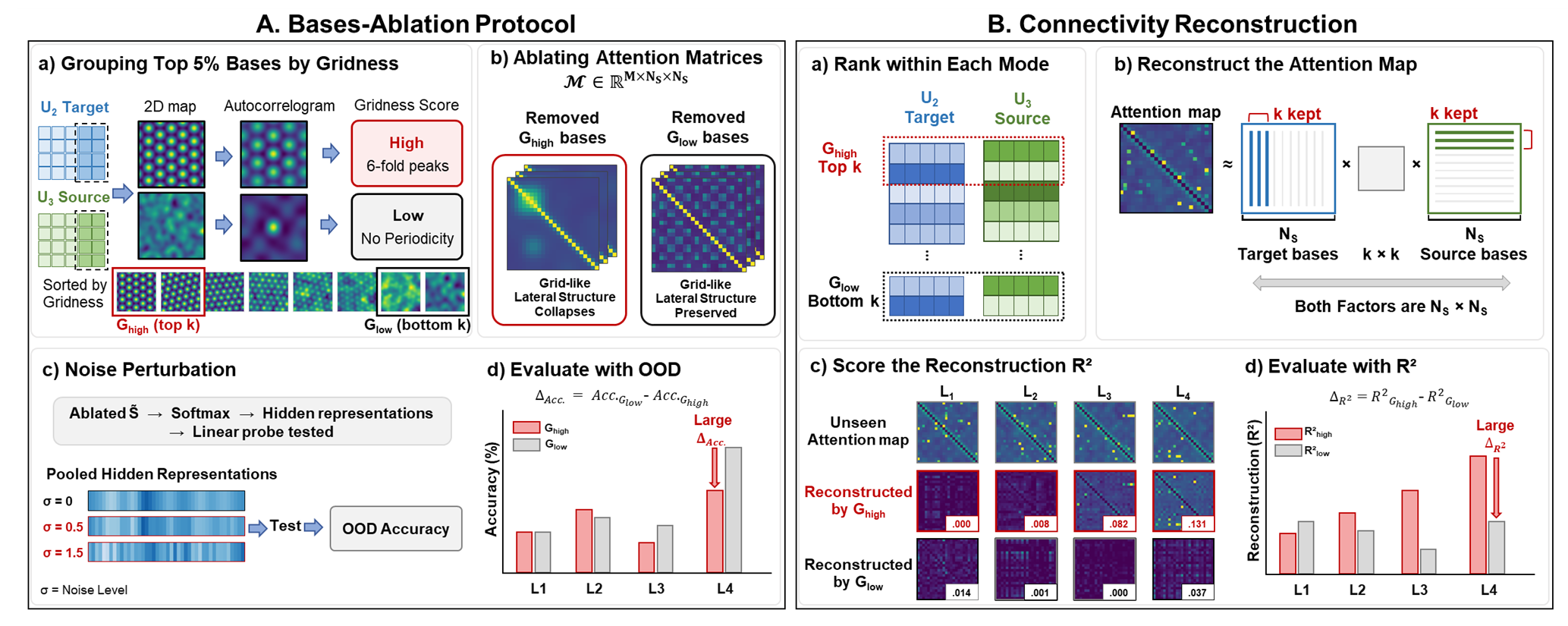}
    \caption{\textbf{Protocols for testing the functional relevance of grid-like bases.} 
    \textbf{A. Bases-ablation Protocol.} 
    a) Top 5\% spatial bases sorted by gridness into $G_{\mathrm{high}}$ and $G_{\mathrm{low}}$. 
    b) Attention matrices ablated by projecting out $G_{\mathrm{high}}$ or $G_{\mathrm{low}}$ subspaces. 
    c) OOD accuracy (Acc.) evaluated under Gaussian noise perturbations. 
    d) OOD Acc. gap, $\Delta_{\mathrm{Acc.}} = \mathrm{Acc.}(G_{\mathrm{low}}) - \mathrm{Acc.}(G_{\mathrm{high}})$, quantifies reliance on $G_{\mathrm{high}}$. 
    \textbf{B. Connectivity Reconstruction.} 
    a) Disjoint top-$k$ ($G_{\mathrm{high}}$) and bottom-$k$ ($G_{\mathrm{low}}$) subsets selected per mode. 
    b) Unseen attention maps reconstructed via bilateral projection. 
    c) Layer-wise attention maps and $R^2$ scores evaluated for each group. 
    d) Reconstruction gap, $\Delta R^2 = R^2_{\mathrm{high}} - R^2_{\mathrm{low}}$, assesses preservation of unseen connectivity.}
    \label{fig:relevance}
\end{figure}

\begin{itemize}
\item \textbf{Appendix~\ref{sec:background}} reviews related dual-stream models in computer vision and computational neuroscience.
\item \textbf{Appendix~\ref{app:sec:graph_framework}} reformulates intra-areal lateral connectivity as graph operators, culminating in Multi-Head Self-Attention as their dynamic instantiation.
\item \textbf{Appendix~\ref{app:sec:gated_recurrence}} describes the temporal dynamics of layer-wise representations via a per-head Gated Recurrent Unit.
\item \textbf{Appendix~\ref{app:sec:alignment}} describes the alignment of dorsal and ventral representation shapes for cross-attention top-down modulation.
\item \textbf{Appendix~\ref{app:extractors}} specifies the pretrained networks for primary dorsal and ventral streams.
\item \textbf{Appendix~\ref{app:classification}} details the classification head: adaptive gated fusion and auxiliary losses.
\item \textbf{Appendix~\ref{sec:bg_grid}} extends the successor representation framework from hippocampal representations to attention matrices.
\item \textbf{Appendix~\ref{app:training}}  details the training protocol, including the selected action classes and hyperparameter settings. 
\item \textbf{Appendix~\ref{app:gate_analysis}} reports the learned per-head $D{\to}V$ gate values $\sigma(\gamma)$.
\item \textbf{Appendix~\ref{app:gridness}} describes gridness analysis pipeline and robustness across retention rates.
\item \textbf{Appendix~\ref{app:baseline_comparison}} compares ReGraph with standard action-recognition baselines (SlowFast, TimeSformer, VideoMAE) in accuracy and in the layer-wise ratio of grid-like bases.
\item \textbf{Appendix~\ref{relevance}} details the bases-ablation and connectivity-reconstruction protocols (Fig.~\ref{fig:relevance}), with sensitivity to the number of ablated bases, noise level, and candidate pool size.

\item \textbf{Appendix~\ref{app:compute}} reports the computational resources and training time.
\item \textbf{Appendix~\ref{app:broader_impact}} discusses broader impacts.
\end{itemize}

All codes and trained model weights are publicly available at \url{https://anonymous.4open.science/r/regraph-351D/README.md}.

\section{Related works}
\label{sec:background}

The dual-stream architecture was first adopted in computer vision to process \emph{appearance} and \emph{motion} for video recognition~\citep{simonyan2014two, feichtenhofer2019slowfast}, and later embraced by computational neuroscience as a model of cortical visual processing. \citet{bakhtiari2021functional} showed that two structurally similar 3D-convolutional pathways trained under a shared self-supervised predictive objective spontaneously develop ventral- and dorsal-like representations aligned with mouse visual cortex. While this offers insight into how functional segregation can naturally emerge from the two-stream architectural constraint alone, the authors did not impose stream-specific biological inductive biases, leaving the computational role of each stream incompletely characterized. \citet{choi2023dual} partly addressed this by implementing retina-inspired M/P input sampling and distinct objectives---spatial attention for the dorsal stream and object recognition for the ventral stream. Although their model clearly demonstrated stream-specific correspondence with human fMRI, thus showing biological validity, most analyses centered on representational alignment rather than the computational principles underlying each stream. To probe the latter, recent work increasingly turns to stream-specific tasks: \citet{han2022modeling} employed spatial relation estimation (\textit{e.g.,} location or orientation prediction), and VeDo-Net~\citep{zareh2024deep} used object-distance estimation, both targeting the dorsal pathway. These models, however, capture only static aspects of dorsal function and largely rely on CNN backbones, leaving room to enrich the architecture with more biologically plausible details.

\section{Modeling spatial lateral connectivity}
\label{app:sec:graph_framework}

\subsection{Formalizing intra-areal lateral connectivity as graph operations}

\paragraph{Roadmap.} We re-express a circuit of intra-areal neuronal ensembles as a graph and describe their neural dynamics through the message-passing paradigm of Graph Neural Networks (GNNs). Among GNN variants, the Graph Convolutional Network (GCN)~\citep{kipf2016semi} provides the simplest and most widely adopted instantiation of message-passing, making it a natural starting point for our formalization.

\paragraph{1) From cortical connectivities to graphs.} The biological principles explained in \S\ref{sec:extended} describe spatially distributed neuronal ensembles communicating through complex, stimulus-driven horizontal pathways. This functional architecture maps naturally onto a graph
$\mathcal{G}=(\mathcal{V},\mathcal{E})$, with the following correspondences:

\begin{itemize}
    \item \textbf{Nodes} $v_i\in\mathcal{V}$: local neuronal ensembles, \textit{e.g.}
    cortical columns each tuned to a specific spatial receptive field.
    \item \textbf{Edges} $(j,i)\in\mathcal{E}$: lateral synaptic pathways
    between neuronal ensembles, each carrying an edge feature $\mathbf{e}_{ji}$ that
    encodes its synaptic efficacy.
    \item \textbf{Adjacency matrix} $\mathbf{A}$: a circuit coupling matrix in which
    the non-zero entry $\mathbf{A}_{ji}$ marks a directed synaptic pathway from
    node $j$ to node $i$.
\end{itemize}

\paragraph{2) Message passing as a mathematical mirror of neural dynamics.} Modeling continuous representations over such structured data is unified under GNN, whose operations
are grounded in the \emph{Message Passing} paradigm~\citep{gilmer2017neural}:
\begin{equation}
\mathbf{h}_{i}^{\ell+1}
\;=\;
\phi\!\left(\,
    \mathbf{h}_{i}^{\ell},\;
    \bigoplus_{j \in \mathcal{N}_i}
        \psi\!\big(\mathbf{h}_{i}^{\ell}, \mathbf{h}_{j}^{\ell}, \mathbf{e}_{ji}\big)
\right),
\label{eq:app_gmp}
\end{equation}
where $\mathbf{h}_{i}^{\ell}$ is the embedding of node $i$ at layer $\ell$,
and $\mathcal{N}_i = \{j \mid \mathbf{e}_{ji}\neq\mathbf{0}\}$ is its
local neighborhood. Each component admits a direct biological interpretation,
summarized in Table~\ref{tab:mp_biology}.

\begin{table}[h]
\centering
\small
\caption{\textbf{Correspondence between message-passing operations and cortical lateral processing.}}
\label{tab:mp_biology}
\begin{tabular}{@{}lll@{}}
\toprule
\textbf{Symbol} & \textbf{GNN role} & \textbf{Biological analog} \\
\midrule
$\psi(\cdot)$    & Message function   & Axonal/synaptic transmission gated by $\mathbf{e}_{ji}$ \\
$\bigoplus$      & Permutation-invariant aggregation & Dendritic integration of postsynaptic potentials \\
$\phi(\cdot)$    & Update function    & Non-linear somatic activation producing the next firing state \\
\bottomrule
\end{tabular}
\end{table}

\paragraph{3) GCN: the simplest-form matrix instantiation.} Among the many instantiations of message passing, GCNs~\citep{kipf2016semi} are the most influential, reducing neighborhood aggregation to a single matrix multiplication followed by a shared linear transformation.
\begin{table}[b]
\centering
\small
\caption{\textbf{GCN viewed as a specific instance of the general message-passing
framework in Eq.~\eqref{eq:app_gmp}.}}
\label{tab:gcn_as_mp}
\begin{tabular}{@{}lll@{}}
\toprule
\textbf{General MP component} & \textbf{General form} & \textbf{GCN realization} \\
\midrule
Message $\psi$        & $\psi(\mathbf{h}_i^{\ell}, \mathbf{h}_j^{\ell}, \mathbf{e}_{ji})$
                      & $\dfrac{1}{\sqrt{\tilde d_i \tilde d_j}}\,\mathbf{W}\mathbf{h}_j^{\ell}$ \\[6pt]
Aggregation $\bigoplus$ & permutation-invariant op
                      & $\displaystyle\sum_{j\in\mathcal{N}_i\cup\{i\}}$ \\[6pt]
Update $\phi$         & $\phi(\mathbf{h}_i^{\ell}, \text{agg})$
                      & $\sigma(\text{agg})$ \\
\bottomrule
\end{tabular}
\end{table}

Self-loops are first added, $\tilde{\mathbf{A}} = \mathbf{A} + \mathbf{I}$,
and the result is symmetrically normalized:
\begin{equation}
\hat{\mathbf{A}} \;=\; \tilde{\mathbf{D}}^{-1/2}\,\tilde{\mathbf{A}}\,\tilde{\mathbf{D}}^{-1/2},
\qquad
\tilde{\mathbf{D}}_{ii} = \sum_j \tilde{\mathbf{A}}_{ij}.
\end{equation}
The normalization weights each incoming message by the inverse geometric mean of the sender and receiver degrees, preventing feature scales from exploding or vanishing under repeated aggregation. The full layer update is then
\begin{equation}
\mathbf{H}^{\ell+1}
\;=\;
\sigma\!\left(
    \hat{\mathbf{A}}\;\mathbf{H}^{\ell}\;\mathbf{W}
\right),
\qquad
\mathbf{H}^{\ell}\in\mathbb{R}^{N\times d},
\label{eq:app_gcn_matrix}
\end{equation}
which decomposes cleanly into three operations:
\begin{itemize}
    \item $\hat{\mathbf{A}}\mathbf{H}^{\ell}$: \emph{spatial aggregation}.
    Row $i$ of $\hat{\mathbf{A}}$ is a vector of normalized connection weights;
    multiplying it against $\mathbf{H}^{\ell}$ yields a weighted sum of node
    $i$'s neighbors. The full matrix product performs this aggregation for all nodes in parallel.
    \item $(\cdot)\mathbf{W}$: \emph{learnable feature transformation} into a new embedding space.
    \item $\sigma(\cdot)$: \emph{non-linear activation}, the analog of neuronal firing.
\end{itemize}
Expanding Eq.~\eqref{eq:app_gcn_matrix} for a single node recovers the
node-centric form
\begin{equation}
\mathbf{h}_{i}^{\ell+1}
\;=\;
\sigma\!\left(
    \sum_{j\in\mathcal{N}_i \cup \{i\}}
        \frac{1}{\sqrt{\tilde{d}_i\,\tilde{d}_j}}\;\mathbf{W}\mathbf{h}_{j}^{\ell}
\right),
\label{eq:app_gcn_node}
\end{equation}
which makes the message-passing structure of GCN explicit. The mapping back to
Eq.~\eqref{eq:app_gmp} is summarized in Table~\ref{tab:gcn_as_mp}.

\subsection{Self-attention as a dynamic graph operator}
\paragraph{Roadmap.} GCN provides an interpretable graph picture of lateral
processing, but its adjacency matrix is \emph{predefined, sparse, and static}, incompatible with at least two cortical properties below: 
\begin{itemize}
    \item \textbf{Hebbian-shaped dense connectivity.} 
    Local cortical ensembles are densely interconnected, and these 
    connections are continuously reshaped by Hebbian-like 
    plasticity---incompatible with a \emph{predefined, sparse} 
    adjacency.
    \item \textbf{Stimulus-driven effective connectivity.} 
    The instantaneous routing through these connections is modulated 
    by current sensory input---incompatible with a \emph{static} 
    adjacency.
\end{itemize}
To reflect these points, we adopted \textbf{Multi-Head Self-Attention (MHSA)}~\citep{vaswani2017attention} as the spatial operator and show below that it is precisely the generalization of GCN satisfying the two cortical properties while preserving the graph-based interpretation.

\paragraph{1) Side-by-side comparison.} The two operators have a highly parallel matrix form:
\begin{equation}
\underbrace{\sigma\!\left(
    \hat{\mathbf{A}}\,\mathbf{H}^{\ell}\,\mathbf{W}
\right)}_{\text{GCN layer}}
\qquad\Longleftrightarrow\qquad
\underbrace{\text{softmax}\!\left(\frac{\mathbf{Q}\mathbf{K}^{\top}}{\sqrt{d_k}}\right)\mathbf{V}}_{\text{Self-Attention}}
\end{equation}
with $\mathbf{Q}=\mathbf{H}^{\ell}\mathbf{W}_Q$,
$\mathbf{K}=\mathbf{H}^{\ell}\mathbf{W}_K$,
$\mathbf{V}=\mathbf{H}^{\ell}\mathbf{W}_V$. Reading them in parallel reveals that self-attention has the same two-stage structure as GCN ---adjacency $\times$ transformed features--- with each replaced by a richer, input-dependent counterpart. The role-by-role correspondence is given in Table~\ref{tab:gcn_vs_attn}.

\begin{table}[h]
\centering
\footnotesize
\caption{\textbf{Component-level equivalence between GCN and self-attention.}
Self-attention preserves the graph-operator structure of GCN but lifts the
adjacency from a static, sparse object to a dynamic, dense one.}
\label{tab:gcn_vs_attn}
\renewcommand{\arraystretch}{1.25}

\begin{tabularx}{\textwidth}{@{} l X X @{}} 
\toprule
\textbf{Functional role}       & \textbf{GCN}                  & \textbf{Self-Attention} \\
\midrule
Adjacency (who connects to whom)  & $\hat{\mathbf{A}}_{ij}$ (scalar, fixed)
                               & $\dfrac{\mathbf{q}_i^{\top}\mathbf{k}_j}{\sqrt{d_k}}$ (input-dependent) \\
Edge-weight normalization      & $1/\sqrt{\tilde d_i \tilde d_j}$ (degree-based)
                               & $\text{softmax}$ over $j$ (similarity-based) \\
Connectivity scope             & Local: $j\in\mathcal{N}_i\cup\{i\}$
                               & Global: every pair $(i,j)$, $1\le i,j\le N$ \\
Source of connectivity         & Predefined graph topology
                               & Learned via $\mathbf{W}_Q,\mathbf{W}_K$ \\
Temporal behavior              & Fixed across all inputs
                               & Recomputed for every $\mathbf{H}^{\ell}$ \\
Transmitted signal             & $\mathbf{W}\mathbf{h}_j^{\ell}$
                               & $\mathbf{V}_j = \mathbf{W}_V\mathbf{h}_j^{\ell}$ \\
Non-linearity                  & Explicit $\sigma(\cdot)$ post-aggregation
                               & Implicit via softmax + downstream MLP \\
\bottomrule
\end{tabularx}
\end{table}

\paragraph{2) Why self-attention satisfies the biological constraints.}
\paragraph{$\mathbf{Q}\mathbf{K}^{\top}$ as a dynamic adjacency matrix.}
In GCN, $\hat{\mathbf{A}}_{ij}$ is a scalar fixed in advance. In
self-attention, this role is played by $[\mathbf{Q}\mathbf{K}^{\top}]_{ij} = \mathbf{q}_i^{\top}\mathbf{k}_j$, the inner product between the $i$-th query and the $j$-th key. Since $\mathbf{q}_i$ and $\mathbf{k}_j$ are linear projections of the current embeddings, this inner product measures the \emph{representational alignment} between the two nodes in the projected space---exactly what an adjacency weight should encode: the strength of the directed edge from $j$ to $i$.

Three properties distinguish this dynamic adjacency from the static
$\hat{\mathbf{A}}$ of GCN:

\begin{itemize}
    \item \textbf{Learned notion of connectivity.} Because
    $\mathbf{W}_Q,\mathbf{W}_K$ are trainable, the network learns
    \emph{what kind of similarity defines a connection}, rather than inheriting
    that decision from a fixed graph topology.
    \item \textbf{Dense connections.} $\mathbf{Q}\mathbf{K}^{\top}$ is computed
    over all $N\times N$ node pairs, yielding a fully connected effective
    adjacency that resembles more an actual local neural circuit.
    \item \textbf{Dynamic across time.} Since $\mathbf{Q}$ and $\mathbf{K}$ are
    functions of $\mathbf{H}^{\ell}$, which itself changes with each new
    sensory input, the adjacency is automatically recomputed per stimulus.
    The same node pair may be strongly connected for one input and weakly
    connected for another --- directly realizing the stimulus-driven
    effective connectivity.
\end{itemize}

\paragraph{$\mathbf{V}$ as the lateral signal content.}
In GCN, $\mathbf{H}^{\ell}\mathbf{W}$ transforms node features prior to
aggregation. In self-attention, the value projection $\mathbf{V}=\mathbf{H}^{\ell}\mathbf{W}_V$ plays an identical role, specifying the content of the lateral signal transmitted along the (now dynamic) edges of the graph. Notably, separating the connectivity computation $(\mathbf{Q},\mathbf{K})$ from the signal computation $(\mathbf{V})$ gives the network independent control over \emph{which neighbors to listen to} and \emph{what to broadcast}---a flexibility GCN lacks, since both are entangled in the single matrix $\mathbf{W}$.

\section{Modeling temporal recurrent connectivity}
\label{app:sec:gated_recurrence}
Within each attention head $h$ at layer $\ell$, the spatially aggregated signal $\tilde{\mathbf{E}}^{\ell}_{h}$ (or its modulated form $\tilde{\mathbf{E}}'^{\,\ell}_{h}$ for the ventral stream; Eq.~\ref{eq:gated_addition}) is integrated with the preceding hidden state $\mathbf{h}_{h}^{\ell}$ via a per-head Gated Recurrent Unit (GRU). The recurrent update unrolls into a gated mixture of the previous hidden state and a newly proposed candidate:

\begin{equation}
\hat{\mathbf{E}}^{\ell}_{h}
\;=\; \mathrm{GRU}\!\left(
    \tilde{\mathbf{E}}^{\ell}_{h},\;
    \mathbf{h}_{h}^{\ell}
\right)
\;=\; (\mathbf{1} - z) \odot \mathbf{h}_{h}^{\ell}
   \;+\; z \odot \tilde{\mathbf{c}}_{h}^{\ell}
\;\;\in\; \mathbb{R}^{B \times N \times d_h},
\label{eq:app_gru}
\end{equation}

where $\odot$ denotes element-wise multiplication, and the gates and candidate are jointly computed from the pair $(\tilde{\mathbf{E}}^{\ell}_{h}, \mathbf{h}_{h}^{\ell})$:

\begin{itemize}
    \item \textbf{Update gate} $\;z = z\!\left(\tilde{\mathbf{E}}^{\ell}_{h},\, \mathbf{h}_{h}^{\ell}\right) \in [0, 1]^{B \times N \times d_h}$: a sigmoid-bounded gate controlling how strongly the new candidate overwrites the prior hidden state. Values close to $0$ preserve past temporal context, while values close to $1$ admit the new candidate---providing a learnable proxy for biological gating that selectively retains or refreshes the local state.

    \item \textbf{Reset gate} $\;r = r\!\left(\tilde{\mathbf{E}}^{\ell}_{h},\, \mathbf{h}_{h}^{\ell}\right) \in [0, 1]^{B \times N \times d_h}$: a second sigmoid-bounded gate that masks the prior hidden state \emph{before} it enters the candidate computation, allowing the unit to selectively forget obsolete past context.

    \item \textbf{Candidate hidden state} $\;\tilde{\mathbf{c}}^{\ell}_{h} = \tilde{\mathbf{c}}^{\ell}_{h}\!\left(r \odot \mathbf{h}_{h}^{\ell},\; \tilde{\mathbf{E}}^{\ell}_{h}\right) \in \mathbb{R}^{B \times N \times d_h}$: a non-linear combination ($\tanh$-activated) of the reset-gated past state and the current spatially aggregated input.
\end{itemize}

Each GRU operates independently per head. This maintains a temporally evolving state for its own functional channel, preserving the multi-channel parallel processing structure (the second biological principle in \S\ref{sec:extended}) established by MHSA across the temporal dimension.

\section{Top-down modulation: dorsal to ventral spatiotemporal alignment}
\label{app:sec:alignment}

To perform the cross-attention top-down modulation described in Section~\ref{sec:topdown}, the dimension of the higher dorsal representation must match that of the ventral one to serve as keys and values. However, the two streams operate on complementary spatiotemporal grids:
\begin{itemize}
    \item \textbf{Dorsal}: $(B,\, T_D,\, N_D,\, d_h) = (B,\, 12,\, 196,\, d_h)$
    \item \textbf{Ventral}: $(B,\, T_V,\, N_V,\, d_h) = (B,\, 3,\, 784,\, d_h)$
\end{itemize}

The M/P scaling factor $\alpha = 4$ ensures the dorsal stream has $\alpha$ times more frames ($T_D = \alpha T_V$, i.e., $12 = 4 \times 3$) and $\alpha$ times fewer spatial tokens ($N_D = N_V / \alpha$, i.e., $196 = 784 / 4$) than the ventral stream. As a result, \emph{moving the temporal axis into the spatial axis}---grouping every $\alpha = 4$ consecutive dorsal frames into a single ventral time step and concatenating their spatial tokens ($4 \times 196 = 784$)---reshapes the dorsal stream representation to exactly match the ventral counterpart, enabling cross-attention between the two streams.

We slice the 12 dorsal frames into 3 non-overlapping windows of 4 consecutive frames and align each window with one of the 3 time steps in the ventral frames. To preserve intra-window temporal order, we add a learnable, zero-initialized positional embedding $\mathbf{p}_\tau \in \mathbb{R}^{d_h}$ ($\tau = 1, \dots, 4$) to each frame; the same $\{\mathbf{p}_1, \dots, \mathbf{p}_4\}$ are reused across all 3 time steps in the ventral frames to encode a consistent slot structure. For each time step $t$ in the ventral frames at layer $\ell$ and head $h$, the four assigned dorsal frames from $\tilde{\mathbf{E}}^{\ell+1}_{h}$ are concatenated along the spatial axis to form the \textbf{dimension-aligned dorsal representation}:
\begin{equation}
\mathbf{C}^{\ell}_{h}[t] \;=\;
\mathrm{Concat}_{\text{spatial}}\!\Big[\,
\tilde{\mathbf{E}}^{\ell+1}_{h}[t_1] + \mathbf{p}_1,\; \dots,\;
\tilde{\mathbf{E}}^{\ell+1}_{h}[t_4] + \mathbf{p}_4
\,\Big]
\;\in\; \mathbb{R}^{B \times 784 \times d_h},
\label{eq:app_dorsal_context}
\end{equation}
where $\{t_1, \dots, t_4\}$ are the four dorsal frames aligned to the time step $t$ in the ventral frames. Stacking across all 3 ventral time steps yields $\mathbf{C}^{\ell}_{h} \in \mathbb{R}^{B \times 3 \times 784 \times d_h}$. 

Cross-attention is then performed using the ventral MHSA output $\tilde{\mathbf{E}}^{\ell}_{h}$ as queries and $\mathbf{C}^{\ell}_{h}$ as both keys and values, yielding the higher-level dorsal context $\tilde{\mathbf{C}}^{\ell}_{h}$ used in Eq.~\ref{eq:gated_addition}:
\begin{equation}
\tilde{\mathbf{C}}^{\ell}_{h} = \mathrm{CrossAttention}(\tilde{\mathbf{E}}^{\ell}_{h},\, \mathbf{C}^{\ell}_{h},\, \mathbf{C}^{\ell}_{h}).
\label{eq:app_cross_attn}
\end{equation}

\section{Stream-specific pretrained networks}
\label{app:extractors}
\S\ref{sec:primary} outlines how raw video frames are asymmetrically sampled into stream-specific inputs ( hereafter as $\mathbf{X}_D$ and $\mathbf{X}_V$). This section details the architectures and selection criteria of the pretrained networks (\textit{i.e.,} \textit{Depth Anything V2, GMFlow, and UniSal}) used to transform these inputs into the final representations $\mathbf{E}_D, \mathbf{E}_V \in \mathbb{R}^{B \times T \times N \times d}$. We leverage task-specific pretrained models to bypass training from scratch and exploit their established functional capabilities. Our model selection follows these criteria:
\begin{enumerate}
    \item \textbf{Stream-specific functional alignment:} Models must be pretrained on tasks that inherently elicit the desired biological properties (\textit{e.g.,} optical flow, depth, and saliency for the M-driven dorsal stream; static object-oriented processing for the P-driven ventral stream).
    \item \textbf{Preservation of 2D spatial layout:} Models must adopt architectures---such as Convolutional Neural Networks (CNNs) or Vision Transformers (ViTs)---whose intermediate layers retain a topographic spatial structure recoverable as a 2D feature map ($C' \times H' \times W'$), ensuring spatial information is maintained rather than collapsed early into a single global vector.
    \item \textbf{Established representational quality:} To ensure rich, domain-specific feature extraction, we select widely recognized models with strong, competitive performance in their respective visual domains.
\end{enumerate}
Following these criteria, both ventral and dorsal streams share a common processing pipeline. Each frame in the input sequence---a single frame for depth and saliency, or a consecutive frame pair (frames at $t$ and $t+1$) for optical flow---passes through the pretrained network, from which we extract intermediate representations. The resulting spatial and channel dimensions are then aligned to meet the predefined ReGraph requirements, that is, the number of spatial tokens ($N_{Dorsal}$ or $N_{Ventral}$) and the embedding dimension $d = 512$.

\subsection{Primary dorsal stream: multi-cue spatiotemporal pretrained networks}
The grayscale sequence $\mathbf{X}_{Dorsal}$ was processed by three pretrained encoder-decoder networks targeting depth, optical flow, and saliency. Since all three tasks produce dense spatial predictions, we extracted intermediate representations from the \emph{decoder} block which projects task semantics back into a 2D layout---late enough to carry task-specific semantics, yet early enough to retain a meaningful channel dimension before the network collapses to its final low-dimensional output (\textit{e.g.,} 1-channel depth map).

\begin{itemize}
    \item \textit{Depth} (Depth Anything V2~\citep{yang2024depth}, License: Apache 2.0): DINOv2-ViT-S/14 encoder + DPT decoder (four fusion blocks: \texttt{refinenet4} $\to$ \texttt{refinenet1}), followed by a head producing a 1-channel scalar depth map. We extracted the representations from \textbf{refinenet2}, where coarse global geometry has been recovered. Output: $64 \times 16 \times 16$.

    \item \textit{Optical flow} (GMFlow~\citep{xu2022gmflow}, License: Apache 2.0): CNN backbone + Transformer global correlation module + spatial-refinement upsampler (Conv $\to$ ReLU $\to$ Conv), followed by a head producing 2-channel flow. We extracted the representations from the upsampler's \textbf{first convolutional layer}, where Transformer correlations are first integrated with local spatial features. Output: $256 \times 28 \times 28$. Since flow estimation requires consecutive frame pairs, this yields $T_D-1$ temporal outputs; we duplicate the first to restore $T_D$.

    \item \textit{Saliency} (UniSal~\citep{droste2020unified}, License: Apache 2.0): MobileNetV2 backbone with temporal recurrent layers + spatial decoder with skip connections (\texttt{upsampling\_1} $\to$ \texttt{upsampling\_2}), followed by a head producing a 1-channel saliency map. We extracted the representations from \textbf{upsampling\_2}, where temporal dynamics from the recurrent layers and spatial priors from the skip connections are jointly integrated. Output: $128 \times 28 \times 28$.

\end{itemize}

\textbf{Spatial alignment and integration.}
The three pretrained outputs differed in spatial size ($16{\times}16$ for Depth, $28{\times}28$ for Optical flow and Saliency) and channel dimension ($C'=64, 256, 128$, respectively), prompting us to spatially downsample each map to the dorsal grid ($14{\times}14$), flatten into $N_D=196$ tokens, and linearly project to $d=512$. This yields three aligned cues $\mathbf{f}^{\text{depth}}, \mathbf{f}^{\text{optical flow}}, \mathbf{f}^{\text{saliency}} \in \mathbb{R}^{B \times T_D \times N_D \times d}$, all matching the target dimension of the primary dorsal representation $\mathbf{E}_D$. We then integrated them via $H=8$-head cross-attention (matching the downstream ReGraph layers), with distinct roles: \emph{saliency} acts as a spatial \emph{router} (queries and keys), while \emph{depth} and \emph{flow} carry the routed \emph{content} (values, concatenated along the channel axis). A residual connection re-injects the saliency feature into the output, following the standard Transformer convention of adding the residual to the query stream. Conceptually, saliency tells the model \emph{where to look}, while depth and optical flow tell it \emph{how to contextualize}. The fusion is:
\begin{equation}
    \mathbf{E}_D = \mathrm{LN}\Big(\mathrm{GELU}\big(
    \mathrm{Attention}(\mathbf{Q}, \mathbf{K}, \mathbf{V}) \mathbf{W}_O
    \big) + \mathbf{f}^{\mathrm{sal}}\Big),
\end{equation}
with $\mathbf{Q} = \mathbf{f}^{\mathrm{sal}}\mathbf{W}_Q$, $\mathbf{K} = \mathbf{f}^{\mathrm{sal}}\mathbf{W}_K$, and $\mathbf{V} = [\mathbf{f}^{\mathrm{depth}} \| \mathbf{f}^{\mathrm{flow}}]\mathbf{W}_V$. Here, $\mathbf{W}_Q, \mathbf{W}_K, \mathbf{W}_O \in \mathbb{R}^{d \times d}$, while $\mathbf{W}_V \in \mathbb{R}^{2d \times d}$ compresses the concatenated depth+flow channels back to $d=512$. Stacking the frame-wise representations along the temporal axis yields the primary dorsal representation $\mathbf{E}_D \in \mathbb{R}^{B \times T_D \times N_D \times d}$ ($T_D=12$, $N_D=196$, $d=512$).

\begin{table}[!h]
\centering
\caption{\textbf{Pretrained network specifications for the primary visual stage.} ``Extraction Block'' denotes the specific layer from which representations are drawn. ``Output Dim.'' specifies the raw channel and spatial shape ($C' \times H' \times W'$) per frame, prior to spatial resampling and linear projection that produce the tokenized form $\mathbb{R}^{N_S \times d}$.}
\label{tab:extractors}
\footnotesize
\setlength{\tabcolsep}{4pt}
\renewcommand{\arraystretch}{1.25}
\begin{tabularx}{\textwidth}{@{} l l l c c >{\raggedright\arraybackslash}X @{}}
\toprule
\textbf{Stream} & \textbf{Cue} & \textbf{Model} & \textbf{Extraction Block} & \textbf{Output Dim.} & \textbf{Post-processing} \\
\midrule
\multirow{3}{*}{Dorsal} 
& Depth    & Depth Anything V2 & \texttt{refinenet2}        & $64 \times 16 \times 16$  & \multirow{3}{=}{Flatten to $N_D{=}196$, project to $d{=}512$, fuse via cross-attention $\to \mathbf{E}_D$} \\
& Flow     & GMFlow            & \texttt{Conv1} (upsampler) & $256 \times 28 \times 28$ & \\
& Saliency & UniSal            & \texttt{upsample\_2}       & $128 \times 28 \times 28$ & \\
\midrule
Ventral 
& Object   & ViT-S/8 (DINO)    & \texttt{Block 9}           & $384 \times 28 \times 28$ & Discard \texttt{[CLS]}, project to $d{=}512 \to \mathbf{E}_V$ ($N_V{=}784$) \\
\bottomrule
\end{tabularx}
\end{table}

\subsection{Primary ventral stream: object-oriented pretrained networks}
For the ventral stream, the RGB sequence $\mathbf{X}_V$ was processed frame-by-frame. We used ViT-Small (License: Apache 2.0)~\citep{dosovitskiy2020image} pretrained on ImageNet via the DINO self-supervised framework~\citep{caron2021emerging}. Unlike the encoder-decoder architectures used in the dorsal stream, ViT consists of 12 Transformer encoder blocks that preserve spatial token count and channel dimension throughout. We extracted intermediate representations from \textbf{Block 9} ($\sim$75\% depth), late enough to abstract object identities. After discarding the \texttt{[CLS]} token, the output formed $N_V = 784$ spatial tokens with $C' = 384$.

\textbf{Spatial alignment.} 
Since the extracted resolution already matches the ventral grid ($N_V = 28 \times 28 = 784$), no spatial resampling is needed. We linearly projected the 784 patch tokens from $C'=384$ to the shared embedding $d=512$. Stacking the frame-wise representations along the temporal axis yields the primary ventral representation $\mathbf{E}_V \in \mathbb{R}^{B \times T_V \times N_V \times d}$ ($T_V=3$, $N_V=784$, $d=512$). 

\section{Classification head and training objective}
\label{app:classification}

\paragraph{Spatiotemporal pooling.}
After $L$ ReGraph layers, the terminal hidden states of each stream, $\hat{\mathbf{E}}^{L}_{Dorsal}$ and $\hat{\mathbf{E}}^{L}_{Ventral}$, are spatiotemporally pooled via global average pooling over both the temporal and spatial axes:
\begin{equation}
\mathbf{g}_{S} = \frac{1}{T_S \cdot N_S} \sum_{t=1}^{T_S} \sum_{n=1}^{N_S}
\hat{\mathbf{E}}^{L}_{S}[t, n] \;\in\; \mathbb{R}^{B \times d},
\quad S \in \{Dorsal, Ventral\}.
\end{equation}
This yields stream-level global representations $\mathbf{g}_{Dorsal}$ and $\mathbf{g}_{Ventral}$ for the dorsal and ventral streams, respectively.

\paragraph{Adaptive gated fusion.}
Classifying fine-grained actions in SSV2 requires a delicate interplay between spatiotemporal motion and detailed object states, but the optimal ratio between the two varies across action classes. For instance, \emph{``Opening something''} versus \emph{``Pretending to open something''} share nearly identical hand motion trajectories (dorsal-stream-oriented); only the object's physical state (ventral-stream-oriented) disambiguates them. Naive concatenation of the two stream representations is therefore sub-optimal.

To address this, we drew inspiration from two paradigms: Mixture of Experts (MoE)~\citep{jacobs1991adaptive}, which treats each stream as an independent expert producing its own class predictions, and dynamic multi-modal fusion~\citep{arevalo2017gated}, which learns instance-specific weights to blend these predictions. The two streams produce independent class logits:
\begin{align}
\hat{\mathbf{y}}_{Dorsal} &= \mathrm{Dropout}(\mathbf{g}_{Dorsal})\mathbf{W}_{Dorsal}, \\
\hat{\mathbf{y}}_{Ventral} &= \mathrm{Dropout}(\mathbf{g}_{Ventral})\mathbf{W}_{Ventral},
\end{align}
where $\hat{\mathbf{y}}_{Dorsal}, \hat{\mathbf{y}}_{Ventral} \in \mathbb{R}^{B \times C}$ ($C$: number of action classes). A per-sample blending weight is then computed from the concatenated stream representations:
\begin{equation}
\boldsymbol{\alpha} = \sigma\!\Big(\mathrm{MLP}\big([\mathbf{g}_{Dorsal} \| \mathbf{g}_{Ventral}]\big)\Big) \;\in\; (0,1)^{B \times C},
\end{equation}
where the MLP is a two-layer ReLU network and $\sigma$ is the sigmoid. The final prediction is the dynamically weighted sum of the two experts:
\begin{equation}
\hat{\mathbf{y}}_{Fused} = \boldsymbol{\alpha} \odot \hat{\mathbf{y}}_{Dorsal} + (1 - \boldsymbol{\alpha}) \odot \hat{\mathbf{y}}_{Ventral}.
\label{eq:gated_ensemble}
\end{equation}
During training, $\boldsymbol{\alpha}$ is softly clamped to $[0.05, 0.95]$ to prevent mode collapse, allowing the model to vary the dorsal-to-ventral ratio per sample---favoring dorsal for motion-heavy actions and ventral when fine-grained object details are critical.

\paragraph{Training objective with auxiliary losses.}
Relying solely on the gating mechanism during training is problematic: since SSV2 consists mostly of motion-heavy actions, the gate quickly learns to rely almost entirely on the dorsal stream, leaving the ventral stream with weak gradients despite its role in capturing object-oriented details for many fine-grained actions.

To address this early imbalance, we adopt the auxiliary loss principle~\citep{szegedy2015going}, attaching an independent classification head to each stream so that both receive strong, direct learning signals. The total training loss is:
\begin{equation}
\mathcal{L} = \mathcal{L}_{\mathrm{CE}}(\hat{\mathbf{y}}_{\mathrm{Fused}},\, y)
\;+\; \lambda_{Dorsal} \cdot \mathcal{L}_{\mathrm{CE}}(\hat{\mathbf{y}}_{\mathrm{Dorsal}},\, y)
\;+\; \lambda_{Ventral} \cdot \mathcal{L}_{\mathrm{CE}}(\hat{\mathbf{y}}_{\mathrm{Ventral}},\, y),
\label{eq:total_loss}
\end{equation}
where $\mathcal{L}_{\mathrm{CE}}$ is standard cross-entropy. We allocate a total auxiliary budget of $0.6$, following common practice~\citep{szegedy2015going}. Within this budget, a grid search over $(\lambda_{Dorsal}, \lambda_{Ventral}) \in \{(0.5, 0.1),\, (0.3, 0.3),\, (0.1, 0.5)\}$ identified $(0.5, 0.1)$ as the best configuration, which we adopted as the final hyperparameter setting (Table~\ref{tab:ablation}). At inference time, the auxiliary heads are discarded and only the adaptively fused prediction $\hat{\mathbf{y}}_{\mathrm{Fused}}$ (Eq.~\ref{eq:gated_ensemble}) is used.

\section{Extending the successor representation: from hippocampal representations to attention matrices}
\label{sec:bg_grid}
A body of computational work has shown that grid-like periodic patterns arise naturally from the low-dimensional structure of hippocampal representations~\citep{dordek2016extracting, sorscher2023unified}. Most pertinent to our approach is the foundational work of \citet{stachenfeld2017hippocampus} who analyzed a successor representation (SR). The SR admits a natural graph-theoretic reading: it can be regarded as a spatial adjacency matrix where nodes represent discrete locations and edges encode expected future occupancies from one location to another. \citet{stachenfeld2017hippocampus} demonstrated that spectrally decomposing the SR yields eigenvectors whose dominant components, by encoding the relative importance of each spatial position (a spatial weighting map), can be interpreted as analogs of grid cell firing-rate maps.

We adopted this analytical framework, which links grid-like patterns to the dominant bases of a spatial adjacency matrix. Crucially, however, \citet{stachenfeld2017hippocampus} focused exclusively on the hippocampal-entorhinal system, without examining whether similar mechanisms might operate upstream. Our core hypothesis extends this principle: we posited that such dominant-structure-driven grid patterns do not abruptly arise within the MEC, but rather develop progressively along the sensory-driven dorsal stream preceding the MEC. To test this, we applied a similar decomposition methodology to the spatial adjacency matrices within our dorsal and ventral streams, explicitly quantifying and comparing the proportion of significant grid-like patterns among the dominant bases that emerge layer by layer. Building on this framework, we treated each layer's pre-softmax attention matrix $\tilde{\mathbf{A}}_{l,h} \propto \mathbf{Q}\mathbf{K}^{\!\top}$ as a direct SR analog---a spatial adjacency matrix whose nodes are spatial tokens (each corresponding to a fixed spatial patch location) and whose edges are connectivity strengths between nodes---and analyzed it following the same procedure. To extract basis structures stable across samples, we stacked the adjacency matrices across \textit{M} samples of each action class into a single tensor $\in \mathbb{R}^{M \times N_s \times N_s}$. We then applied Higher-Order Singular Value Decomposition (HOSVD)~\citep{de2000multilinear}, which jointly factorizes all matrices along a shared spatial mode. Reshaping each resulting basis into its 2D spatial layout yielded a spatial weighting map indicating the relative importance of each spatial position.

\section{Training details}
\label{app:training}

\subsection{Selected action classes}
\label{app:classes}

\begin{figure}[!htbp]
    \centering
    \includegraphics[width=\textwidth]{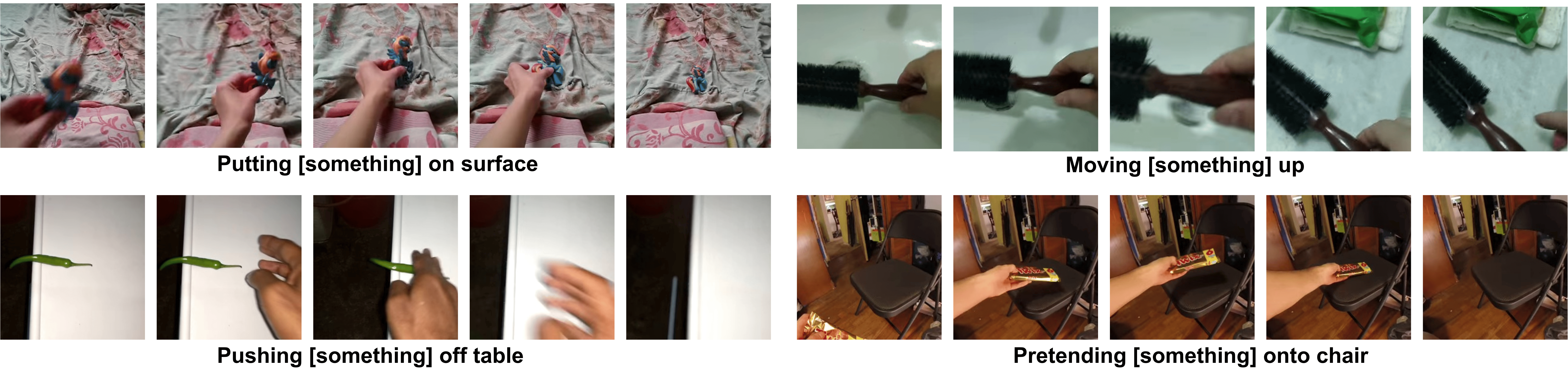}
    \caption{\textbf{Example videos and corresponding templated descriptions~\citep{goyal2017something}.}}
    \label{fig:ssv2_examples}
\end{figure}

To construct a balanced subset of SSV2~\citep{goyal2017something} (used under the Qualcomm AI Research Data License Agreement, research-use only, non-commercial) in terms of training-test sample size, we retained the 33 classes with at least 1{,}000 training and 200 test samples, then uniformly subsampled each class to exactly these counts (Table~\ref{tab:classes}). Figure~\ref{fig:ssv2_examples} shows representative video examples and their corresponding templated descriptions.

\begin{table}[htbp]
\centering
\caption{\textbf{The 33 action classes selected from SSV2, sorted by class ID.} IDs correspond to the original 174-class indexing of the SSV2 dataset.}
\label{tab:classes}
\small
\begin{tabular}{c l}
\toprule
\textbf{ID} & \textbf{Action Class} \\
\midrule
6   & Covering something with something \\
12  & Dropping something onto something \\
15  & Hitting something with something \\
16  & Holding something \\
19  & Holding something next to something \\
36  & Moving something and something away from each other \\
37  & Moving something and something closer to each other \\
45  & Moving something up \\
46  & Opening something \\
47  & Picking something up \\
49  & Plugging something into something \\
56  & Poking something so lightly that it doesn't or almost doesn't move \\
67  & Pretending to open something without actually opening it \\
93  & Pushing something from left to right \\
94  & Pushing something from right to left \\
99  & Pushing something so that it falls off the table \\
100 & Pushing something so that it slightly moves \\
109 & Putting something on a surface \\
114 & Putting something similar to other things that are already on the table \\
122 & Rolling something on a flat surface \\
125 & Showing something behind something \\
129 & Showing that something is empty \\
134 & Something falling like a feather or paper \\
135 & Something falling like a rock \\
140 & Spinning something that quickly stops spinning \\
143 & Squeezing something \\
144 & Stacking number of something \\
145 & Stuffing something into something \\
148 & Taking something out of something \\
149 & Tearing something into two pieces \\
150 & Tearing something just a little bit \\
164 & Turning something upside down \\
171 & Uncovering something \\
\bottomrule
\end{tabular}
\end{table}

\subsection{Hyperparameter settings}
\label{app:hparams}

All models were trained with AdamW and cosine learning rate scheduling with linear warmup. Early stopping was applied based on validation accuracy with a patience of 8 epochs. Table~\ref{tab:hparams} summarizes the settings.

Single-stream references (Dorsal-Only and Ventral-Only) shared the same training configuration. All dual-stream variants had substantially more parameters than the single-stream references, requiring stronger regularization: we increased weight decay (0.03 $\to$ 0.05) and label smoothing (0.1 $\to$ 0.15), and additionally applied Mixup/CutMix augmentation. Gradient accumulation steps were also raised (4 $\to$ 6) to yield a larger effective batch size (30 vs.\ 20).

\begin{table}[htbp]
\centering
\caption{\textbf{Training hyperparameters.} The "Single-Stream" column 
applies to both single-stream references (Dorsal-Only and 
Ventral-Only), and the "Dual-Stream" column applies to all 
dual-stream variants.}
\label{tab:hparams}
\footnotesize
\renewcommand{\arraystretch}{1.1}
\setlength{\tabcolsep}{4pt}
\begin{tabular*}{0.7\textwidth}{@{\extracolsep{\fill}} l cc @{}}
\toprule
\textbf{Hyperparameter} & \textbf{Single-Stream} & \textbf{Dual-Stream} \\
\midrule
Learning rate (ReGraph layers)           & 5e-4  & 3e-4  \\
Learning rate (pretrained networks)       & 5e-5    & 3e-5    \\
Weight decay            & 0.03  & 0.05  \\
Batch size              & 5     & 5     \\
Gradient accumulation   & 4     & 6     \\
Effective batch size    & 20    & 30    \\
Epochs                  & 100   & 100   \\
Warmup epochs           & 3     & 5     \\
Early stopping patience & 8     & 8     \\
Label smoothing         & 0.1   & 0.15  \\
Mixup / CutMix $\alpha$ & -- / -- & 0.3 / 0.3 \\
Mix probability         & --    & 0.5   \\
\bottomrule
\end{tabular*}
\end{table}

\section{Head-wise \texorpdfstring{$D \to V$}{D -> V} gate values}
\label{app:gate_analysis}

Table~\ref{tab:gate_values} reports the per-head gate values across the extended ventral stream hierarchy ($V_k$, $k \in \{1, 2, 3\}$). The gate value $\sigma(\gamma)$ stays near initialization at the early ventral layers ($V_1$, $V_2$) but opens sharply at $V_3$ across all heads.

\begin{table}[htbp]
\centering
\caption{\textbf{Learned $D{\to}V$ gate values $\sigma(\gamma)$ per head at each ventral layer of \emph{ReGraph}.} $V_k$ ($k \in \{1, 2, 3\}$) denotes the $k$-th ventral layer; $V_4$ is omitted as it receives no $D{\to}V$ modulation. Bold highlights the sharp opening at $V_3$ across all heads.}
\label{tab:gate_values}
\footnotesize
\renewcommand{\arraystretch}{1.1}
\setlength{\tabcolsep}{4pt}
\begin{tabular*}{\textwidth}{@{\extracolsep{\fill}}l cccccccc c@{}}
\toprule
\textbf{Layer} & \textbf{H1} & \textbf{H2} & \textbf{H3} & \textbf{H4} & \textbf{H5} & \textbf{H6} & \textbf{H7} & \textbf{H8} & \textbf{Mean ($\pm$ std)} \\
\midrule
$V_1$ & 0.292 & 0.283 & 0.249 & 0.294 & 0.326 & 0.262 & 0.253 & 0.269 & 0.279 ($\pm$0.024) \\
$V_2$ & 0.277 & 0.287 & 0.313 & 0.298 & 0.277 & 0.283 & 0.290 & 0.283 & 0.289 ($\pm$0.011) \\
$V_3$ & \textbf{0.444} & \textbf{0.479} & \textbf{0.478} & \textbf{0.491} & \textbf{0.481} & \textbf{0.444} & \textbf{0.482} & \textbf{0.472} & \textbf{0.471 ($\pm$0.017)} \\
\bottomrule
\end{tabular*}
\end{table}

\section{Gridness analysis details}
\label{app:gridness}
\subsection{Computational procedure}
\label{app:gridness_procedure}
To determine whether each layer's lateral connectivity contains hexagonal grid-like spatial bases, we followed three steps: \textit{i)} extract dominant spatial bases from the pre-softmax attention matrices via HOSVD, \textit{ii)} compute the gridness score for each basis, and \textit{iii) }test each score against a per-basis null distribution.

\textbf{Adjacency matrix stacking and HOSVD.}
We restricted the analysis to the 13 \emph{qualifying} action classes with above-median per-class accuracy, collecting 30 correctly classified training samples per class. At each layer $l$ and head $h$, the per-sample pre-softmax attention matrix $\tilde{\mathbf{A}}_{l,h}^{(S)} \in \mathbb{R}^{N_S \times N_S}$ (where $S \in \{\mathrm{Dorsal}, \mathrm{Ventral}\}$) was extracted with its diagonal zeroed, and stacked across the 30 samples into a tensor $\boldsymbol{\mathcal{T}} \in \mathbb{R}^{30 \times N_S \times N_S}$ with axes (Sample $\times$ Target $\times$ Source), where the Target and Source modes correspond to the row and column dimensions of the attention matrix---representing the receiving and sending neuronal ensembles of each lateral connection, respectively. We then applied HOSVD~\citep{de2000multilinear} only to the two spatial axes (Target, Source)---\emph{not} to the Sample axis---so that the extracted bases reflect connectivity structure that is invariant across samples within the class. This produces two orthonormal basis sets:
\begin{itemize}
    \item $\mathbf{u}_k^{(\mathrm{target})}$ --- \emph{incoming} connectivity bases (Target mode),
    \item $\mathbf{u}_k^{(\mathrm{source})}$ --- \emph{outgoing} connectivity bases (Source mode),
\end{itemize}
each indexed $k = 1, \dots, N_S$ and accompanied by its singular value.

\textbf{Top-$p\%$ basis retention.}
Within each stream and layer, we sorted the bases of each mode by singular-value magnitude in descending order and retain the top $p\%$ ($p \in \{3, 5, 7\}$). The number of bases retained per class per layer is:
\begin{equation}
K_p^{(\text{class})} = \underbrace{H}_{8\,\text{heads}} \times \underbrace{2}_{\text{tgt+src}} \times \underbrace{\lfloor p\% \cdot N_S \rfloor}_{\text{number of top-}p\%\text{ bases per mode}},
\end{equation}
totaling $K_p = K_p^{(\text{class})} \times 13$ when aggregated across the 13 classes. 
At $p = 5\%$, the total number of retained bases per layer is:
\begin{itemize}
    \item Dorsal ($N = 196$): $K_5 = 1{,}872$
    \item Ventral ($N = 784$): $K_5 = 8{,}112$
\end{itemize}

\textbf{Gridness scoring.}
Each retained spatial basis was reshaped to its 2D spatial layout ($14 \times 14$ for dorsal, $28 \times 28$ for ventral), serving as a spatial weighting map analogous to a grid cell firing-rate map (Appendix~\ref{sec:bg_grid}). From this map, a gridness score was computed following the standard procedure of~\citet{sargolini2006conjunctive}: forming the 2D spatial autocorrelogram, identifying the six peaks closest to the center, and quantifying the six-fold rotational symmetry of this hexagonal arrangement.

\textbf{Per-basis significance test.}
Following~\citet{banino2018vector}, we constructed a null distribution \emph{per basis} by computing gridness scores on 100 spatially shuffled copies of that spatial basis map, and set the basis-specific threshold $\theta_{\text{null}}$ at the 95th percentile. A basis was classified as a significant grid-like spatial basis only if its score exceeds \emph{its own} $\theta_{\text{null}}$. Because the null distribution of gridness scores depends on the grid resolution ($14 \times 14$ vs.\ $28 \times 28$), this per-basis shuffling automatically yields resolution-adjusted thresholds, enabling fair comparison across streams.

\subsection{Detailed gridness results (Top-5\% retention rate)}
\label{app:gridness_5pct}
Tables~\ref{tab:detail5_counts} and~\ref{tab:detail5_scores} provide detailed layer-wise results extending Table~\ref{tab:gridness_main} in the main text. Following the procedure in Appendix~\ref{app:gridness_procedure}, a basis was classified as a \emph{grid-like spatial basis} only if its gridness score exceeds its own 95th-percentile null threshold.  

Table~\ref{tab:detail5_counts} reports the absolute count of these significant grid-like spatial bases ($n_{\text{grid}}$) and their proportion relative to the total number of retained bases ($K$). Table~\ref{tab:detail5_scores} reports the mean significance threshold ($\bar{\theta}_{\text{null}}$) and the mean gridness score ($\bar{g}_{\text{sig}}$). Crucially, note that $\bar{\theta}_{\text{null}}$ is computed over the full pool of $K$ retained bases, whereas $\bar{g}_{\text{sig}}$ is computed exclusively over the subset of $n_{\text{grid}}$ bases that passed the significance threshold. At the top-5\% retention rate, the total number of bases per layer is $K = 1{,}872$ for the dorsal stream and $K = 8{,}112$ for ventral self-attention.
\begin{table}[!h]
\centering
\caption{\textbf{Counts and ratios of grid-like spatial bases (Top-5\% retention rate).} Each cell shows $n_{\text{grid}}$ (\textit{Ratio\%}), aggregated across classes and heads. Bold highlights ReGraph's dorsal stream, exhibiting monotonic emergence of grid-like bases.}
\label{tab:detail5_counts}
\footnotesize
\renewcommand{\arraystretch}{1.1}
\setlength{\tabcolsep}{4pt}
\begin{tabular*}{\textwidth}{@{\extracolsep{\fill}} l l ccccc @{}}
\toprule
\textbf{Model} & \textbf{Stream} & \textbf{L1} & \textbf{L2} & \textbf{L3} & \textbf{L4} & \textbf{Avg} \\
\midrule
Dorsal-Only & Dorsal  & 0 (0.0\%) & 105 (5.6\%) & 75 (4.0\%) & 48 (2.6\%) & 228 (3.0\%) \\
\midrule
\multirow{2}{*}{Unmodulated}
  & Dorsal  & 1 (0.1\%) & 161 (8.6\%) & 74 (4.0\%) & 118 (6.3\%) & 354 (4.7\%) \\
  & Ventral & 9 (0.1\%) & 553 (6.8\%) & 479 (5.9\%) & 478 (5.9\%) & 1519 (4.7\%) \\
\midrule
\multirow{2}{*}{ReGraph}
  & Dorsal  & \textbf{0 (0.0\%)} & \textbf{95 (5.1\%)} & \textbf{164 (8.8\%)} & \textbf{286 (15.3\%)} & \textbf{545 (7.3\%)} \\
  & Ventral & 9 (0.1\%) & 525 (6.5\%) & 483 (6.0\%) & 588 (7.2\%) & 1605 (4.9\%) \\
\bottomrule
\end{tabular*}
\end{table}

\begin{table}[!h]
\centering
\caption{\textbf{Mean significance thresholds and gridness scores (Top-5\% retention rate).} Each cell shows $\bar{\theta}_{\text{null}}\,/\,\bar{g}_{\text{sig}}$; ``--'' indicates no grid-like bases were identified. Bold highlights ReGraph's dorsal stream.}
\label{tab:detail5_scores}
\footnotesize
\renewcommand{\arraystretch}{1.1}
\setlength{\tabcolsep}{4pt}
\begin{tabular*}{\textwidth}{@{\extracolsep{\fill}} l l ccccc @{}}
\toprule
\textbf{Model} & \textbf{Stream} & \textbf{L1} & \textbf{L2} & \textbf{L3} & \textbf{L4} & \textbf{Avg} \\
\midrule
Dorsal-Only & Dorsal & .244\,/\,-- & .240\,/\,.272 & .244\,/\,.324 & .242\,/\,.341 & .243\,/\,.312 \\
\midrule
\multirow{2}{*}{Unmodulated}
  & Dorsal  & .242\,/\,.269 & .240\,/\,.409 & .242\,/\,.390 & .241\,/\,.382 & .241\,/\,.363 \\
  & Ventral & .216\,/\,.274 & .216\,/\,.340 & .198\,/\,.301 & .186\,/\,.294 & .204\,/\,.302 \\
\midrule
\multirow{2}{*}{ReGraph}
  & Dorsal  & \textbf{.243\,/\,--} & \textbf{.242\,/\,.366} & \textbf{.245\,/\,.390} & \textbf{.244\,/\,.351} & \textbf{.244\,/\,.369} \\
  & Ventral & .216\,/\,.276 & .216\,/\,.340 & .193\,/\,.313 & .178\,/\,.283 & .201\,/\,.303 \\
\bottomrule
\end{tabular*}
\end{table}

\begin{table}[!h]
\centering
\caption{\textbf{Ratio of grid-like bases (\%) in the ReGraph dorsal stream across retention rates.} The monotonic increase across layers is consistently reproduced at all rates.}
\label{tab:gridness_robustness}
\footnotesize
\renewcommand{\arraystretch}{1.1}
\setlength{\tabcolsep}{4pt}
\begin{tabular*}{0.5\textwidth}{@{\extracolsep{\fill}}l ccc@{}}
\toprule
\textbf{Layer} & \textbf{3\%} & \textbf{5\%} & \textbf{7\%} \\
\midrule
L1 & 0.0  & 0.0  & 0.2  \\
L2 & 1.4  & 5.1  & 5.6  \\
L3 & 12.1 & 8.8  & 8.3  \\
L4 & 20.2 & 15.3 & 12.1 \\
\bottomrule
\end{tabular*}
\end{table}

\subsection{Robustness across retention rates}
\label{app:gridness_robustness}
To ensure that the monotonic increase in grid-like spatial bases along the ReGraph dorsal stream is not an artifact of the 5\% retention rate, we repeated the analysis at 3\% and 7\%. Table~\ref{tab:gridness_robustness} reports the ratio of grid-like spatial bases in the ReGraph dorsal stream across these rates; the monotonic pattern is consistently reproduced.


\begin{figure}[tb]
    \centering
    \includegraphics[width=\textwidth]{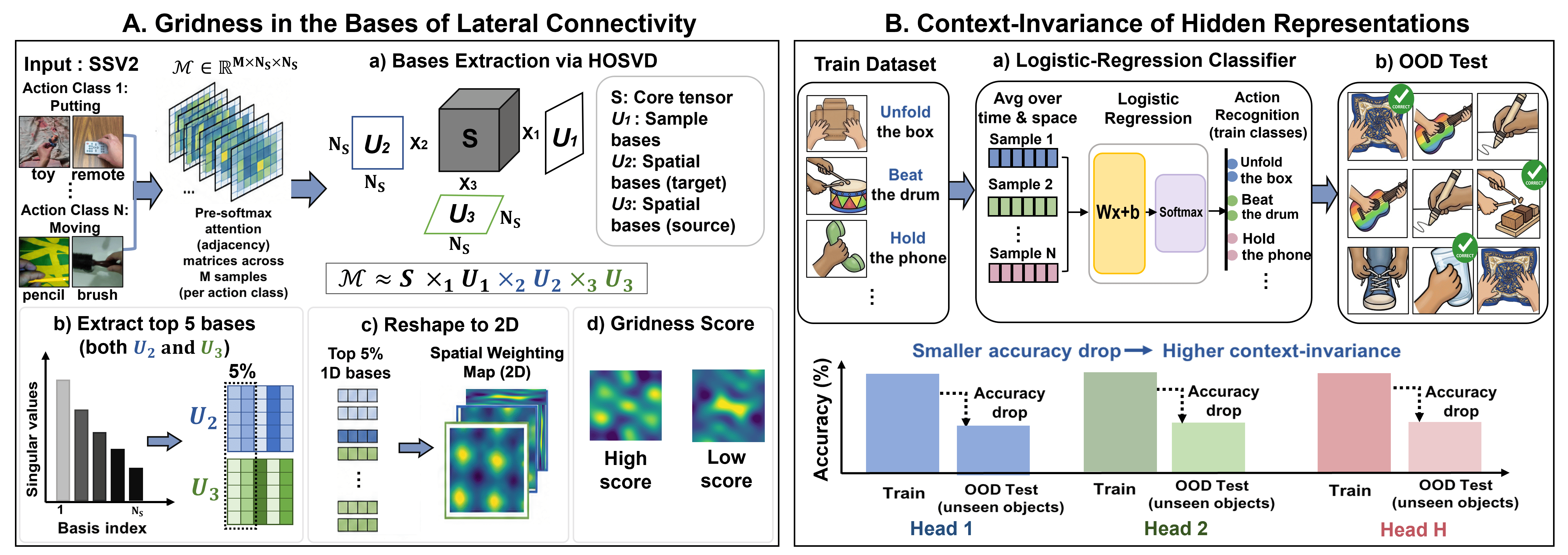}
    \caption{\textbf{RPA pipeline.} 
    \textbf{A.} a) SSV2 attention matrices ($\mathcal{M}$) decomposed via HOSVD to isolate spatial bases ($U_2, U_3$). b-c) Top 3, 5, 7\% of bases (ranked by singular value) reshaped into 2D spatial weighting maps. d) Gridness scores computed per map. \textbf{B.} a) Logistic-regression classifier trained on spatiotemporally pooled representations. b) Evaluation on the test set (unseen objects) assesses context-invariance; smaller accuracy drop indicates higher invariance.}
    \label{fig:RPA}
\end{figure}

\section{Comparison with standard action-recognition baselines}
\label{app:baseline_comparison}
\paragraph{Protocol.} All baselines were trained and evaluated under a protocol
identical to ReGraph's: the same 33-class SSV2 subset and train/validation
split, the same balanced class sampling (seed 42), the same early-stopping
rule, and the same 6-view ensemble evaluation on the test set. Each baseline was initialized from its
official public checkpoint and fine-tuned on our SSV2 subset, using the training recipe published in its original paper (optimizer, learning-rate schedule, input length, and normalization
statistics). Note that all three baselines entered this comparison as end-to-end models pretrained on a large action-recognition dataset (Kinetics-400), whereas ReGraph's pretrained front-end was derived from heterogeneous datasets across task-agnostic vision tasks (depth, optical flow, saliency, and object recognition), and its backend (33.6M-parameter) was trained from random initialization on SSV2 alone.

\begin{table}[!h]
\centering
\caption{\textbf{Action-recognition comparison on the 33-class SSV2 subset.}
Top-1 accuracy (\%) under the shared 6-view evaluation protocol.}
\label{tab:baseline_acc}
\footnotesize
\renewcommand{\arraystretch}{1.1}
\setlength{\tabcolsep}{4pt}
\begin{tabular*}{\textwidth}{@{\extracolsep{\fill}} l l l c @{}}
\toprule
\textbf{Model} & \textbf{Initialization} & \textbf{Fine-tuning} & \textbf{Top-1} \\
\midrule
VideoMAE       & K400 (self-supervised)                 & \multirow{4}{*}{SSV2-33 subset} & 80.4 \\
\textbf{ReGraph (ours)} & \textbf{depth/flow/saliency/object} & & \textbf{74.6} \\
SlowFast       & K400 (supervised)                      & & 66.1 \\
TimeSformer    & ImageNet-21K $\rightarrow$ K400 (supervised) & & 62.3 \\
\bottomrule
\end{tabular*}
\end{table}

\paragraph{Grid-like representations in the baselines.} We further applied the
RPA (Appendix~\ref{app:gridness_procedure}) to the three baselines. VideoMAE and TimeSformer both build on a 12-layer ViT encoder backbone whose
public checkpoints were pretrained on $224\times224$ inputs with $16\times16$
patches, so each frame is natively tokenized into the same $14\times14$
spatial grid (196 tokens) as ReGraph's dorsal stream; their
per-frame $196\times196$ pre-softmax attention scores therefore entered the RPA
exactly as ReGraph's did. We sampled every third encoder layer (3/6/9/12,
denoted $L_1$--$L_4$) to match ReGraph's four-layer depth axis. SlowFast is a convolutional network and thus has no attention scores to
analyze. Its backbone comprises five convolutional blocks; we took the feature maps of the first four (denoted
$L_1$--$L_4$) and average-pooled each onto a common $14\times14$ grid, matching
ReGraph's spatial resolution (the dorsal side), while the last block was excluded as its
$7\times7$ feature map was coarser than this grid. The $196\times196$
cosine-similarity matrix among the 196 pooled positions was then used in place
of the attention score matrix, separately for the fast and slow pathways. All subsequent
steps---per-class HOSVD, gridness scoring, and the 95th-percentile null
threshold---were identical to the ReGraph analysis. As shown in
Table~\ref{tab:baseline_gridness}, despite their competitive accuracy, none of
the baselines exhibits the progressive layer-wise emergence of grid-like
patterns observed in ReGraph's dorsal stream.

\begin{table}[!h]
\centering
\caption{\textbf{Ratio of grid-like bases (\%) in baseline models.}}
\label{tab:baseline_gridness}
\footnotesize
\renewcommand{\arraystretch}{1.1}
\setlength{\tabcolsep}{4pt}
\begin{tabular*}{0.72\textwidth}{@{\extracolsep{\fill}} l cccc @{}}
\toprule
\textbf{Model} & \textbf{L1} & \textbf{L2} & \textbf{L3} & \textbf{L4} \\
\midrule
\textbf{ReGraph --- dorsal} & \textbf{0.0} & \textbf{5.1} & \textbf{8.8} & \textbf{15.3} \\
ReGraph --- ventral & 0.1 & 6.5 & 6.0 & 7.2 \\
VideoMAE            & 2.4 & 1.6 & 1.0 & 0.9 \\
TimeSformer         & 1.1 & 0.9 & 1.0 & 1.4 \\
SlowFast --- fast   & 1.6 & 1.6 & 0.0 & 0.0 \\
SlowFast --- slow   & 1.6 & 0.0 & 0.0 & 0.0 \\
\bottomrule
\end{tabular*}
\end{table}
\color{black}
\section{Detailed protocols for bases-ablation and connectivity reconstruction analyses}
\label{relevance}

This appendix provides detailed protocols for the Bases-Ablation and Connectivity Reconstruction analyses introduced in the main text (Table~\ref{tab:relevance}). Both analyses were conducted on ReGraph's extended dorsal stream, reusing the same training and OOD test samples as well as the per-class, per-layer, and per-head HOSVD basis extraction procedure described in the RPA protocol (Appendix~\ref{app:gridness_procedure}).

\subsection{Bases-ablation analysis protocol}

The primary goal of this analysis is to causally isolate the specific contribution of grid-like bases to OOD generalization. To achieve this, we selectively removed the connectivity subspace corresponding to either the most grid-like ($G_{\mathrm{high}}$) or the least grid-like ($G_{\mathrm{low}}$) bases during the inference forward pass. The complete procedure consists of the following steps:

\textbf{1. Basis Grouping.} We first constructed a joint candidate ``\textit{base pool}'', comprising the top 5\% of target-mode and source-mode bases ranked by their singular values. We then sorted this combined pool by gridness in descending order, assigning the top $k$ bases to $G_{\mathrm{high}}$ and the bottom $k$ bases to $G_{\mathrm{low}}$. While we conducted this ablation across $k \in \{1, 5, 9\}$, we focused primarily on the $k=9$ setting in the main text, as the diverging functional trends between $G_{\mathrm{high}}$ and $G_{\mathrm{low}}$ are most realized at this scale (detailed comparisons for all $k$ are provided in Table~\ref{tab:app_ablation_k}).

\textbf{2. Projection Operators ($\mathbf{P}$).} To isolate the functional footprint of a given group $G_g$ ($g \in \{\mathrm{high}, \mathrm{low}\}$), we define projection operators that capture the spatial communication subspaces spanned by the target and source bases from $G_g$:
\begin{equation}
\mathbf{P}_{\mathrm{target}}^{(g)} = \mathbf{Q}_{\mathrm{target}}^{(g)} (\mathbf{Q}_{\mathrm{target}}^{(g)})^{\top}, \qquad \mathbf{P}_{\mathrm{source}}^{(g)} = \mathbf{Q}_{\mathrm{source}}^{(g)} (\mathbf{Q}_{\mathrm{source}}^{(g)})^{\top},
\end{equation}
where $\mathbf{Q}_{\mathrm{target}}^{(g)}$ and $\mathbf{Q}_{\mathrm{source}}^{(g)}$ are orthonormal column matrices containing the respective basis vectors of $G_g$. Intuitively, each $\mathbf{P}$ acts as a subspace identifier: it extracts the lateral connectivity features that exclusively align with the selected grid-like ($G_{\mathrm{high}}$) or non-grid-like ($G_{\mathrm{low}}$) patterns.

\textbf{3. Attention Score Ablation.} At every timestep of the inference forward pass, we nullified the selected subspace from the lateral connectivity by applying the orthogonal complement ($\mathbf{I}-\mathbf{P}$) to the pre-softmax attention score $\mathbf{S}$:
\begin{equation}
\tilde{\mathbf{S}} = \mathrm{Diag}(\mathbf{S}) + \mathrm{off}\Big((\mathbf{I} - \mathbf{P}_{\mathrm{target}}^{(g)}) \, \mathrm{off}(\mathbf{S}) \, (\mathbf{I} - \mathbf{P}_{\mathrm{source}}^{(g)})\Big),
\end{equation}
where $\mathrm{off}(\cdot)$ zeroes the diagonal elements and $\mathrm{Diag}(\cdot)$ preserves them. Multiplying $(\mathbf{I} - \mathbf{P})$ bilaterally serves as a two-sided structural filter on the routing matrix:
\begin{itemize}
\item Right-multiplying by $(\mathbf{I} - \mathbf{P}_{\mathrm{source}}^{(g)})$ filters out outgoing signals originating from the selected source subspace.
\item Left-multiplying by $(\mathbf{I} - \mathbf{P}_{\mathrm{target}}^{(g)})$ strips incoming signals that would otherwise be received into the selected target subspace.
\end{itemize}
Consequently, this operation completely severs any cross-token interaction channels mediated by the chosen bases. 

\textbf{4. Evaluation.} Using the ablated attention score $\tilde{\mathbf{S}}$, the model proceeded with the softmax and subsequent updates to produce modified hidden representations. These representations were then used as training and testing samples to fit and evaluate an OOD linear probe. Crucially, because the spatial bases were estimated per class while a test video's class had to remain strictly unknown at inference, the projection operator derived from one specific class was applied uniformly to all test videos. Repeating this procedure for each of the 13 classes yielded 13 independent runs per condition (i.e., ablating $G_{\mathrm{high}}$ or $G_{\mathrm{low}}$), from which we report the mean OOD performance.

\begin{table}[!h]
\centering
\caption{\textbf{OOD performance to the number of ablated bases ($k$).} Cells report OOD accuracy under $G_{\mathrm{high}}$ / \ $G_{\mathrm{low}}$ ablation [gap $\Delta_{Acc.}\!\uparrow$]. A larger positive $\Delta_{Acc.}$ indicates greater reliance on $G_{\mathrm{high}}$. \textbf{Bold} highlights the prominent reliance surge at higher layers ($L_3$--$L_4$), which is most pronounced at $k=9$.}
\label{tab:app_ablation_k}
\footnotesize
\renewcommand{\arraystretch}{1.1}
\setlength{\tabcolsep}{4pt}
\begin{tabular*}{\textwidth}{@{\extracolsep{\fill}} l ccc @{}}
\toprule
\multirow{2}{*}{\textbf{Layer}} & \multicolumn{3}{c}{\textbf{Number of Ablated Bases ($k$)}} \\
\cmidrule(lr){2-4}
 & $k{=}1$ : $G_{\text{high}}$/$G_{\text{low}}$ $[\Delta_{Acc.}\!\uparrow]$
 & $k{=}5$ : $G_{\text{high}}$/$G_{\text{low}}$ $[\Delta_{Acc.}\!\uparrow]$
 & $k{=}9$ : $G_{\text{high}}$/$G_{\text{low}}$ $[\Delta_{Acc.}\!\uparrow]$ \\
\midrule
$L_1$ & 82.0 / 81.9 [$-$0.1] & 81.9 / 81.6 [$-$0.3] & 82.0 / 81.7 [$-$0.3] \\
$L_2$ & 81.3 / 81.5 [$+$0.2] & 80.7 / 81.9 [$+$1.1] & 82.6 / 82.9 [$+$0.3] \\
$L_3$ & \textbf{76.0 / 78.3 [$+$2.4]} & \textbf{69.7 / 77.5 [$+$7.8]} & \textbf{70.3 / 77.3 [$+$7.0]} \\
$L_4$ & \textbf{97.2 / 97.4 [$+$0.2]} & \textbf{93.6 / 96.6 [$+$3.0]} & \textbf{89.1 / 95.1 [$+$6.0]} \\
\bottomrule
\end{tabular*}
\end{table}

\paragraph{Noise perturbation.} For the noise-perturbation results reported in the main text (Table~\ref{tab:relevance}), we applied Gaussian noise to the updated hidden representations of \emph{ReGraph} after they were processed through the $k=9$ ablated lateral connectivity. Specifically, independent Gaussian noise was added to every dimension of these representations, with its magnitude scaled to a noise level $\sigma$ times that dimension's standard deviation across the training representations. These noise-perturbed final hidden representations were then used as testing samples to evaluate the linear probe.

\subsection{Connectivity-reconstruction analysis protocol}

To investigate a possible structural mechanism underlying the greater reliance on $G_{\mathrm{high}}$ to construct lateral connectivity patterns, we tested whether high-gridness bases extracted from training samples also captured more of the attention structure of unseen OOD samples.

\paragraph{Selection of high- and low-gridness bases.} Unlike the Bases-Ablation analysis, which evaluates OOD performance using the remaining bases after removing a targeted set $G_{\mathrm{g}}$, the reconstruction analysis assesses how accurately $G_{\mathrm{g}}$ itself can recover the original attention matrix of unseen samples. Because the reconstructed matrix is formed by the outer product of target- and source-mode bases, omitting either mode results in a zero matrix. To prevent such reconstruction failure from a degenerate zero matrix, we therefore ranked candidates by gridness within each mode and extracted subsets of size $k \in \{1, 5, 9\}$ for both $G_{\mathrm{high}}$ and $G_{\mathrm{low}}$ per mode. Because $G_{\mathrm{high}}$ and $G_{\mathrm{low}}$ must be disjoint, evaluating the maximum subset size ($k=9$) requires 18 distinct candidates per mode. The default 5\% pool provides only 9 candidates, falling short of this requirement. To secure sufficient non-overlapping candidates, we expanded the pool to the leading 30\%, 40\%, and 50\% of singular directions in each mode. (Note that results for $k=1$ are omitted from Table~\ref{tab:app_recon_pool} due to negligible reconstruction scores.)

\paragraph{Reconstruction and interpretation of $R^2$.}
For each class, layer, and head, let $\mathbf{A}_{\mathrm{unseen}}$ denote a diagonal-zeroed pre-softmax attention matrix from an unseen OOD video. To evaluate whether the spatial bases extracted from the training samples could generalize to unseen ones, we reconstructed this attention matrix $\mathbf{A}_{\mathrm{unseen}}$ using the selected bases $\mathbf{Q}_\mathrm{target}^{(g)}, \mathbf{Q}_\mathrm{source}^{(g)} \in \mathbb{R}^{N \times k}$ extracted from the train samples (where $g \in \{\mathrm{high}, \mathrm{low}\}$). The reconstructed matrix $\widehat{\mathbf{A}}^{(g)}$ was computed as:
\begin{equation}
    \widehat{\mathbf{A}}^{(g)} = \mathbf{Q}_\mathrm{target}^{(g)} \left[ \bigl(\mathbf{Q}_\mathrm{target}^{(g)}\bigr)^\top \mathbf{A}_{\mathrm{unseen}} \mathbf{Q}_\mathrm{source}^{(g)} \right] \bigl(\mathbf{Q}_\mathrm{source}^{(g)}\bigr)^\top.
\end{equation}
This operation projects the unseen matrix onto the subspaces spanned by the selected training bases ($\mathbf{Q}_\mathrm{target}^{(g)}$ and $\mathbf{Q}_\mathrm{source}^{(g)}$). Effectively, it discards all unselected connectivity patterns, preserving only the structural components that can be represented by the chosen subspace. We then normalized the reconstruction error by $\|\mathbf{A}_{\mathrm{unseen}}\|_F^2$ to calculate the reconstruction score $R_g^2$:
\begin{equation}
    R_g^2 = \frac{\left\| \bigl(\mathbf{Q}_\mathrm{target}^{(g)}\bigr)^\top \mathbf{A}_{\mathrm{unseen}} \mathbf{Q}_\mathrm{source}^{(g)} \right\|_F^2}{\|\mathbf{A}_{\mathrm{unseen}}\|_F^2} = 1 - \frac{\|\mathbf{A}_{\mathrm{unseen}} - \widehat{\mathbf{A}}^{(g)}\|_F^2}{\|\mathbf{A}_{\mathrm{unseen}}\|_F^2}.
\end{equation}
As intuitively shown in the rightmost term, a smaller discrepancy between the original unseen matrix and its reconstruction yields a higher $R_\mathrm{g}^2$ score. In other words, a higher $R_\mathrm{g}^2$ directly indicates that the selected bases $\mathbf{Q}$ extracted from the training data are effectively utilized to represent the core connectivity pattern of unseen OOD samples.

\begin{table}[!h]
\centering
\caption{\textbf{Reconstruction $R^2$ across pool sizes and subset sizes.} Each cell reports reconstruction scores under $G_{\mathrm{high}}$ / $G_{\mathrm{low}}$ [gap $\Delta_{R^2} = R^2_{G_{\mathrm{high}}} - R^2_{G_{\mathrm{low}}}\uparrow$], averaged over $13 \times 8$ (class, head) pairs ($k{=}1$ omitted as $R^2 < 0.01$). The bottom row reports the layer-wise mean of the gaps ($\Delta_{R^2}$). \textbf{Bold} highlights the prominent surge at $L_3$ and peak at $L_4$ on average across configurations.}
\label{tab:app_recon_pool}
\footnotesize
\setlength{\tabcolsep}{4pt}
\resizebox{\columnwidth}{!}{%
\begin{tabular}{l l cccc}
\toprule
\textbf{Pool} & $k$ & \textbf{$L_1$} & \textbf{$L_2$} & \textbf{$L_3$} & \textbf{$L_4$} \\
\midrule
30\% & 5 & .005 / .000 [$+$0.005] & .001 / .003 [$-$.002] & .055 / .000 [\textbf{$+$.055}] & .105 / .022 [\textbf{$+$.084}] \\
30\% & 9 & .025 / .000 [$+$0.025] & .003 / .066 [$-$.062] & .120 / .001 [\textbf{$+$.118}] & .153 / .052 [\textbf{$+$.101}] \\
40\% & 5 & .004 / .000 [$+$0.004] & .001 / .000 [$+$0.001] & .048 / .000 [\textbf{$+$.048}] & .084 / .016 [\textbf{$+$.068}] \\
40\% & 9 & .014 / .000 [$+$0.014] & .001 / .008 [$-$.007] & \textbf{.082 / .000 [$+$0.082]} & \textbf{.131 / .037 [$+$0.094]} \\
50\% & 5 & .004 / .000 [$+$0.004] & .000 / .000 [$+$0.000] & .044 / .000 [\textbf{$+$.044}] & .060 / .009 [\textbf{$+$.051}] \\
50\% & 9 & .012 / .000 [$+$0.012] & .001 / .000 [$+$0.001] & .058 / .000 [\textbf{$+$.058}] & .123 / .022 [\textbf{$+$.101}] \\
\midrule
\multicolumn{2}{l}{\textbf{Layer Mean ($\Delta$)}} & [$+$0.010] & [$-$.000] & \textbf{[$+$0.066]} & \textbf{[$+$0.082]} \\
\bottomrule
\end{tabular}%
}
\end{table}

\paragraph{Detailed reconstruction results.}
To maintain consistency with the Bases-Ablation setting, we evaluated the reconstruction performance ($R^2$) across the same subset sizes ($k \in \{1, 5, 9\}$) matching the ablation analysis and varying candidate pools (30\%, 40\%, and 50\%). (Note that results for $k=1$ are omitted from Table~\ref{tab:app_recon_pool} due to near-zero scores across all layers). The results demonstrate a clear and consistent layer-wise trajectory: as shown in the layer-wise means (bottom row of Table~\ref{tab:app_recon_pool}), the reconstruction gap is negligible in early layers ($L_1, L_2$), sharply surges at $L_3$, and peaks at $L_4$ on average across configurations. To clearly illustrate this consistent trend in the main text, we report the representative setting where this contrast is most pronounced ($40\%$ pool with $k=9$).


\section{Computational resources}
\label{app:compute}

All experiments were conducted on a single workstation with 2$\times$ NVIDIA
RTX 5090 GPUs (32\,GB each), 256\,GB system RAM, and a
2\,TB NVMe SSD for dataset and checkpoint storage. All models are trained
in PyTorch~2.0+ using DistributedDataParallel across the two GPUs.
Table~\ref{tab:training_time} reports the per-variant training time and
the total compute for the seven reported ablations.

\begin{table}[htbp]
\centering
\caption{\textbf{Compute resources and wall-clock training time for each
reported ablation.} All models trained on 2$\times$ NVIDIA RTX 5090 GPUs
(32\,GB each), with maximum 100 epochs and early stopping at patience$=$8.
Wall-clock includes both training and the final 6-view ensemble evaluation;
GPU-hours $=$ wall-clock $\times$ 2.}
\label{tab:training_time}
\footnotesize
\renewcommand{\arraystretch}{1.15}
\setlength{\tabcolsep}{6pt}
\begin{tabular*}{\textwidth}{@{\extracolsep{\fill}} l c cccc @{}}
\toprule
\textbf{Variant} & \textbf{Top-1 (\%)} & \textbf{Epochs} & \textbf{h/epoch} & \textbf{Wall-clock} & \textbf{GPU-hours} \\
\midrule
\multicolumn{6}{l}{\textit{Single-stream references}} \\
Dorsal-Only                              & 71.69 & 24 & 1.7 & $\sim$43\,h & $\sim$86  \\
Ventral-Only                             & 60.38 & 15 & 0.2 & $\sim$4\,h  & $\sim$8   \\
\midrule
\multicolumn{6}{l}{\textit{Dorsal--Ventral biological trait asymmetry}} \\
(a) Trait-Symmetric                      & 69.78 & 22 & 0.8 & $\sim$19\,h & $\sim$38  \\
(b) Dorsal-trait Only                    & 72.42 & 29 & 1.3 & $\sim$38\,h & $\sim$76  \\
(c) Ventral-trait Only                   & 70.43 & 21 & 0.8 & $\sim$18\,h & $\sim$36  \\
(d) \textbf{Fully Asymmetric (ReGraph)}  & \textbf{74.57} & \textbf{27} & \textbf{1.8} & \textbf{$\sim$50\,h} & \textbf{$\sim$100} \\
\midrule
\multicolumn{6}{l}{\textit{$D{\to}V$ modulation removal}} \\
Unmodulated                              & 72.25 & 28 & 1.8 & $\sim$53\,h & $\sim$106 \\
\midrule
\multicolumn{4}{l}{\textbf{Total compute (the seven reported ablations)}} & \textbf{$\sim$225\,h} & \textbf{$\sim$450} \\
\bottomrule
\end{tabular*}
\end{table}

\paragraph{Storage.}
The Something-Something V2 dataset (videos plus pre-extracted frames at
$224 \times 224$) requires approximately 500\,GB of disk space. Each
trained checkpoint is approximately 360\,MB (\texttt{best\_model\_ema.pth});
the seven released checkpoints total $\sim$2.0\,GB and are hosted on
Zenodo (DOI~\href{https://doi.org/10.5281/zenodo.19917046}{10.5281/zenodo.19917046}).

\paragraph{Computational scalability and limitations.}
\label{app:computational_limit}
While the full ReGraph model can be trained on our SSV2 subset within approximately 50 hours on a standard workstation (Table~\ref{tab:training_time}), the architecture possesses inherent computational limitations regarding scalability. The intra-areal lateral connectivity is implemented via MHSA (\S\ref{sec:extended}), which scales quadratically ($O(N^2)$) with the number of spatial tokens. Although this cost is carefully bounded in our current design by restricting the spatial grids to $N_D=196$ and $N_V=784$, extending this framework to high-resolution videos without heavy downsampling would incur prohibitive memory and compute overhead. Furthermore, expanding the temporal window to reason over long-duration events would linearly increase the recurrent update costs ($O(T)$). Addressing these bottlenecks---potentially through sparse or linear attention mechanisms---remains a necessary step for scaling this biologically inspired architecture to more complex, unconstrained visual domains.

Additionally, these intensive computational demands constrained the scope of our evaluation. To thoroughly explore the architectural design space—which required training multiple dual-stream variants and single-stream references—we were limited to evaluating each reported configuration with a single random seed. Consequently, while the performance gap is substantial, the statistical variance of the overall Top-1 accuracy could not be fully quantified. Future work utilizing scaled compute resources will conduct multi-seed evaluations to rigorously establish the statistical robustness of these findings.

\paragraph{Preliminary and exploratory experiments.}
The seven ablations reported above represent the final controlled comparison;
the broader research process required additional compute that is \emph{not}
reflected in Table~\ref{tab:training_time}. Specifically, exploratory
experiments---including hyperparameter sweeps (learning rate, weight decay,
gradient accumulation, auxiliary-loss weighting), and early architectural
prototypes (alternative fusion modes, lateral wiring schemes, GRU vs.\
vanilla recurrence)---contributed
an additional $\sim$30--50\% wall-clock time on top of the reported total
($\sim$70--100 GPU-hours). These exploratory results were used only for
internal validation and are not part of any reported claim.

\section{Broader impacts}
\label{app:broader_impact}
The primary motivation of this research is foundational: to elucidate the mechanisms underlying the emergence of relational structures in the brain through artificial intelligence. By modeling the dual visual streams, our ReGraph framework provides a promising computational hypothesis for this phenomenon. Consequently, we anticipate that these findings will offer profound insights into the biological basis of human generalization and catalyze a wide range of subsequent neuroscientific research. 

However, because this capability is driven by optimizing action recognition performance---specifically modeling fine-grained hand--object interactions---we must acknowledge the dual-use nature of this technology. The very ability to reliably abstract ``what action is being performed'' regardless of the background context or specific object instantiation carries inherent privacy risks. If deployed outside of controlled experimental settings, such technologies could be repurposed for pervasive surveillance systems, invasive worker monitoring, or other forms of privacy infringement. To partially mitigate these risks, our public release is limited to research-purpose code and checkpoints trained on a constrained 33-class hand--object interaction subset of SSV2, rather than a deployment-ready surveillance pipeline.

\end{document}